%% file: acl_latex.tex
\pdfoutput=1
\documentclass[11pt]{article}

\usepackage[final]{acl}

\usepackage{times}
\usepackage{latexsym}

\usepackage[T1]{fontenc}
\usepackage[utf8]{inputenc}

\usepackage{microtype}

\usepackage{inconsolata}

\usepackage{graphicx}

\usepackage{color}
\usepackage[dvipsnames, svgnames]{xcolor}
\usepackage{colortbl}
\definecolor{darkblue}{rgb}{0, 0, 0.5}
\definecolor{darkgreen}{rgb}{0, 0.5, 0}
\definecolor{mygreen}{rgb}{0,0.8,0}
\definecolor{myred}{rgb}{0.8,0,0}
\definecolor{mygain}{HTML}{009900}
\definecolor{myloss}{HTML}{FF3300}

\usepackage{CJKutf8} %这个宏包是CJK宏包增强？
\usepackage{microtype}
\usepackage{graphicx}
\usepackage{subcaption}
\usepackage{booktabs}
\usepackage{amsmath}
\usepackage{amssymb}
\usepackage{mathtools}
\usepackage{amsthm}
\usepackage[capitalize,noabbrev]{cleveref}

\usepackage{tabularx}       % 支持自动换行的表格列 (必要)
\usepackage[table]{xcolor}  % 支持单元格颜色 (必要, 用于高亮)
\usepackage{booktabs}       % 提供更美观的表格线条 (推荐)
\usepackage{lineno}
\usepackage[utf8]{inputenc}
\usepackage[T1]{fontenc}
\usepackage{url}
\usepackage{amsfonts}
\usepackage{nicefrac}
\usepackage{latexsym}
\usepackage{array}
\usepackage{multirow}
\usepackage{alltt}
\usepackage{enumerate}
\usepackage{bbm}
\usepackage{eucal}
\usepackage{xspace}
\usepackage{stmaryrd}
\usepackage{bussproofs}
\usepackage{mathpartir}
\usepackage{pgfmath}
\usepackage{xurl}
\usepackage{mdframed}
\usepackage{enumitem}
\usepackage{listings}
\usepackage{xparse}
\usepackage{xifthen}
\usepackage{multicol}
\usepackage{xfrac}
\usepackage{appendix}
\usepackage{etoolbox}
\usepackage{floatflt}
\usepackage{wrapfig}
\usepackage{tablefootnote}
\usepackage{makecell}
\usepackage{verbatim}
\usepackage{setspace}
\usepackage{floatrow}
\usepackage{adjustbox}
\usepackage{cleveref}
\usepackage{pifont}
\usepackage[table]{xcolor}
\usepackage{dashrule}
\usepackage{arydshln}
\usepackage{lipsum}
\usepackage{cleveref}
\usepackage{tcolorbox}
\usepackage{listings}
\tcbuselibrary{listingsutf8} 
\usepackage{CJKutf8}  % 支持 UTF-8 编码的中文

\usepackage{pgfplots}
\usepgfplotslibrary{groupplots}
\usepackage{caption}

\usepackage{subcaption}
\usepackage{geometry}
\usepackage{tikz}
\usetikzlibrary{patterns}
\pgfplotsset{compat=1.17}
\usetikzlibrary{patterns}
\usetikzlibrary{backgrounds}
\usepackage{amssymb}
\usepackage{times}
\usepackage{url}
\usepackage{latexsym}
\usepackage{longtable}

\usepackage{graphicx} 
\usepackage{adjustbox}
\usepackage{booktabs}
\usepackage{natbib}
\usepackage{multirow}

\usepackage{pgfplots}
\usepackage{times}
\usepackage{arydshln}

\usetikzlibrary{arrows.meta, positioning, fit, shapes.geometric, arrows}

\usepackage{ltxtable}

\usepackage{titlesec}
\titlespacing*{\paragraph}{0pt}{0.5em}{0.5em}  % 调整段落前后间距

\usepackage{makecell}
\usepackage{hyperref}
\usepackage{algorithm}
\usepackage[%
    noEnd=true,%
    indLines=true,%
    italicComments=false,%
    commentColor=darkgreen,%
]{algpseudocodex}

\title{NiuTrans.LMT: Toward Inclusive and Scalable Multilingual Machine Translation with LLMs}

\author{
    \textbf{Yingfeng Luo\textsuperscript{1}\thanks{\xspace Equal contribution.}},
    \textbf{Ziqiang Xu\textsuperscript{1}\footnotemark[1]}, 
    \textbf{Yuxuan Ouyang\textsuperscript{1}}, 
    \textbf{Murun Yang\textsuperscript{1}},
    \textbf{Dingyang Lin\textsuperscript{1}} \\ 
    \textbf{Kaiyan Chang\textsuperscript{1}}, 
    \textbf{Tong Zheng\textsuperscript{1}},
    \textbf{Bei Li\textsuperscript{1}}, 
    \textbf{Peinan Feng\textsuperscript{1}},
    \textbf{Quan Du\textsuperscript{2}},
    \textbf{Tong Xiao\textsuperscript{1,2}\thanks{\xspace Corresponding author.}},
    \textbf{Jingbo Zhu\textsuperscript{1,2}}
    \\
    \textsuperscript{1} School of Computer Science and Engineering, Northeastern University, Shenyang, China\\
	\textsuperscript{2} NiuTrans Research, Shenyang, China\\
    \texttt{luoyingfeng\_neu@outlook.com}\\
	\texttt{\{xiaotong,zhujingbo\}@mail.neu.edu.cn}
}

\usepackage{ltxtable}
\begin{document}
\begin{CJK*}{UTF8}{gbsn}  % 启动中文支持，字体为 gbsn（宋体）
\maketitle
\begin{abstract}
    \input{sections/0_abstract}

\end{abstract}

\input{sections/1_introduction}

\input{sections/3_method}

\input{sections/4_results}
\input{sections/2_related_work}
\input{sections/5_conclusion}

%\input{sections/4_results_and_analysis}
% \fi
%\bibliographystyle{acl_natbib}
\bibliography{references}

% \newpage\onecolumn
% \addtocontents{toc}{\protect\setcounter{tocdepth}{1}}
% \tableofcontents\newpage

{
\appendices
% \onecolumn
% \begin{@onecolumnfalse}
% \begin{appendices}
%\input{appendices/a_lora}
%\input{appendices/b_data}
\input{appendices/a_setup}

%\input{appendices/d_results}
%\input{appendices/e_prompt}
% \end{appendices}
% \end{@onecolumnfalse}
}
\end{CJK*}
\end{document}

%% file: sections/0_abstract.tex
Large language models have significantly advanced Multilingual Machine Translation (MMT), yet scaling to many languages while keeping quality robust across directions remains challenging.
In this paper, we identify a failure mode of multilingual supervised fine-tuning (SFT) on multi-way parallel data: when such data are reused symmetrically around a pivot language (e.g., English), performance on reverse directions (X $\to$ pivot) can drop substantially.
We term this phenomenon Directional Degeneration and attribute it to excessive many-to-one mappings, which encourage shortcut learning.
We propose Strategic Downsampling (SD), a simple yet effective method to mitigate this degeneration.
In addition, we introduce Parallel Multilingual Prompting (PMP), which augments translation instructions with an auxiliary parallel sentence to promote cross-lingual transfer during training and enables optional test-time enhancement when auxiliary translations are available. 
We further develop \textbf{NiuTrans.LMT} (\textbf{L}arge-scale \textbf{M}ultilingual \textbf{T}ranslation, abbreviated as \textbf{LMT}), a Chinese–English-centric suite of multilingual translation models spanning four sizes (0.6B/1.7B/4B/8B) and covering 60 languages and 234 directions.
Comprehensive evaluations show that LMT is competitive among open-source MMT systems, and that our 4B LMT model performs on par with or better than substantially larger baselines. 
We release our models and project resources to support inclusive and scalable MMT.

\vspace{.3em}
\hspace{.5em}
\includegraphics[height=1.5em]{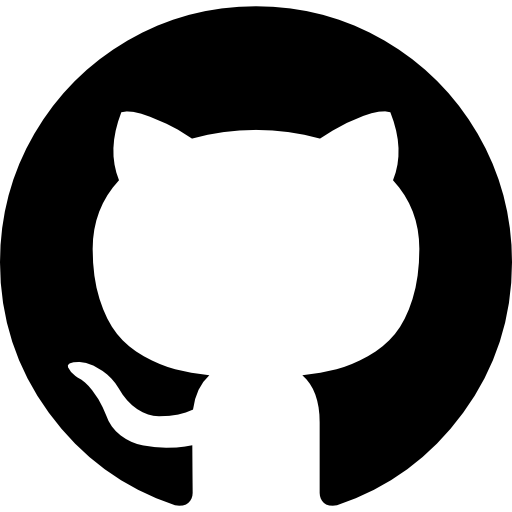}{\hspace{.75em}\parbox{\dimexpr\linewidth-2\fboxsep-2\fboxrule}{\vspace{-5pt} \href{https://github.com/NiuTrans/LMT}{NiuTrans/LMT}}}

%% file: sections/1_introduction.tex
\section{Introduction}
Large language models (LLMs) have reshaped the way we build machine translation (MT) systems. 
Instead of training a dedicated neural MT model from scratch, a widely used approach is to adapt a foundation LLM to translation through post-training \citep{DBLP:journals/corr/abs-2305-18098,DBLP:journals/corr/abs-2402-17733,DBLP:conf/iclr/XuMKHEK25,DBLP:conf/naacl/CuiGLLW25,luoyf2025lamate,DBLP:journals/corr/abs-2502-11223}. 
This shift has substantially improved translation quality and expanded the capability frontier of MT systems.
However, it also raises a key problem for multilingual MT (MMT): how can we adapt foundation LLMs for massively multilingual scale while maintaining robust performance across all translation directions?

\begin{table}[!t]
  % \begin{center}
    \begin{adjustbox}{max width=1\linewidth,center=\linewidth}
    \setlength{\tabcolsep}{3pt}
    \begin{tabular}{lcccc}
      \toprule
      \multirow{2}{*}{\textbf{LLM for MT}} & \multirow{2}{*}{\textbf{\#CPT}} & \multirow{2}{*}{\textbf{\#Langs}}  & \multirow{2}{*}{\makecell{\textbf{Zh-}\\\textbf{centric}}} & \multirow{2}{*}{\makecell{\textbf{Base} \\ \textbf{Model}}}\\
      &&&& \\
      \midrule   % \hline
      BigTranslate \citep{DBLP:journals/corr/abs-2305-18098} & 90B & 102 & \ding{55} & LLaMA \\
      ALMA \citep{DBLP:conf/iclr/Xu0SA24} & 20B  & 6 & \ding{55} & LLaMA-2 \\
      TowerInstruct \citep{DBLP:journals/corr/abs-2402-17733} & 20B & 10 & \ding{55} & LLaMA-2 \\
      \citet{DBLP:conf/naacl/GuoYLWSC24} & 120G & 3 & \ding{55} & LLaMA-2 \\
      X-ALMA \citep{DBLP:conf/iclr/XuMKHEK25} & 40B & 50 & \ding{55} & LLaMA-2 \\
      GemmaX2 \citep{DBLP:conf/naacl/CuiGLLW25} & 54B & 28 & \checkmark & Gemma-2 \\
      Hunyuan-MT \citep{zheng2025hunyuanmt} & - & 33 & \checkmark & Hunyuan-7B \\
      Seed-X \citep{DBLP:journals/corr/abs-2507-13618} & 200B & 28 & \checkmark & - \\
      \hdashline
      \specialrule{0em}{0.2em}{0em}
      \textbf{
      LMT (Ours)} & 90B & 60 & \checkmark & Qwen-3 \\
      \vspace{-0.55cm} \\
      \bottomrule
    \end{tabular}
    \end{adjustbox}
    \caption{Comparison of typical LLM-based MMT models. We summarize their total number of continued pretraining (CPT) data, supported languages, support for Zh-centric translation, and the base models used.}
    \label{tab:mmt_comprasion}
  % \end{center}
\end{table}

Most recent LLM-based MMT systems (representative examples are summarized in \cref{tab:mmt_comprasion}) follow a multi-stage training recipe. 
To bridge the low-resource gap evident in foundation models (see \cref{fig:lang_disparity}, top), the first stage, Continued Pre-training (CPT), serves as the primary step to enhance multilingual competence through large-scale training.
The second stage is supervised fine-tuning (SFT), which aligns the model to high-quality instruction-style translation. 
Since SFT benefits most from clean and diverse supervision \citep{DBLP:conf/nips/ZhouLX0SMMEYYZG23,DBLP:conf/emnlp/ZhuCZH0K24}, it is typically built from human-translated corpora. 
However, for many low-resource directions, such supervision is limited. 
As a result, multi-way human-translated corpora, such as FLORES-200 \cite{DBLP:journals/corr/abs-2207-04672} and NTREX-128 \cite{federmann-etal-2022-ntrex} -like datasets, have become an important source for scaling SFT coverage, since their multi-way structure can theoretically support any directions from a relatively small amount of annotation.

In this work, our findings indicate that current multi-way SFT practices fail when scaling to many languages.
When multi-way corpora are reused symmetrically around a pivot language (e.g., English), we observe an asymmetric outcome: while pivot $\to$ X improves as expected, the reverse X $\to$ pivot directions drop substantially, producing fluent but less faithful translations. 
We term this phenomenon \textbf{Directional Degeneration}. 
We analyze its cause as a data-usage issue. 
In multi-way corpora, the same pivot-language sentence may repeatedly appear as the target for many different sources, which increases target-side repetition and encourages shortcut learning.
To mitigate this, we propose \textbf{Strategic Downsampling (SD)}, a simple data-level strategy that retains full supervision for pivot $\to$ X directions while keeping only a small fraction of reverse-direction instances. 
This change reduces excessive many-to-one mapping and stabilizes reverse-direction translation.

In addition to using multilingual supervision properly, we also study how to elicit cross-lingual transfer more explicitly in MMT. 
We introduce \textbf{Parallel Multilingual Prompting (PMP)}, which augments the translation instruction with an auxiliary parallel sentence as in-context guidance, to facilitate cross-lingual transfer in MMT. 
PMP is applied during training, and it can be activated at inference time as a lightweight enhancement when auxiliary translations are available, including those generated by the model itself.

We instantiate these ideas in \textbf{NiuTrans.LMT} (abbreviated as \textbf{LMT}), a Chinese–English-centric suite of \textbf{L}arge-scale \textbf{M}ultilingual machine \textbf{T}ranslation models covering 60 languages and 234 translation directions, with four model sizes (0.6B/1.7B/4B/8B).
We first perform large-scale CPT on about 90B tokens to strengthen the multilingual base.  
To address the scarcity of Chinese-centric resources (as shown in Figure~\ref{fig:lang_disparity}, bottom), we expand coverage through broad data collection and curation.
We then perform SFT, where SD and PMP are integrated to improve directional robustness and cross-lingual transfer.  
Finally, we apply preference optimization with GRPO \citep{DBLP:journals/corr/abs-2402-03300}, reusing the same supervised pairs from SFT to further refine translation quality.
Comprehensive evaluation shows that LMT is competitive among open-source MMT systems with comparable language coverage.
In particular, LMT-60-4B is on par with or better than substantially larger baselines such as X-ALMA-13B \citep{DBLP:conf/iclr/XuMKHEK25}, Aya-101-13B \citep{2024aya}, and NLLB-54B \citep{DBLP:journals/corr/abs-2207-04672} on the overlapping directions.

In summary, our contributions are threefold:
\begin{itemize}
    \item We identify Directional Degeneration as a failure mode in large-scale multilingual SFT with symmetric multi-way data reuse, analyze its cause, and propose Strategic Downsampling as an effective data-level mitigation.
    \item We introduce Parallel Multilingual Prompting (PMP), which strengthens cross-lingual transfer during training and can be leveraged for optional test-time enhancement.
    \item We present and release LMT, a suite of Chinese–English-centric multilingual translation models in four sizes,  providing broad language coverage and strong performance.
\end{itemize}

\setlength{\fboxsep}{1pt} 
\definecolor{background_color1}{RGB}{255,229,228}
\definecolor{background_color2}{RGB}{227,229,252}
\definecolor{background_color3}{RGB}{228,242,227}

\begin{figure}
\centering
\begin{subfigure}[]{\linewidth}
	\centering
	\includegraphics[width=\linewidth]{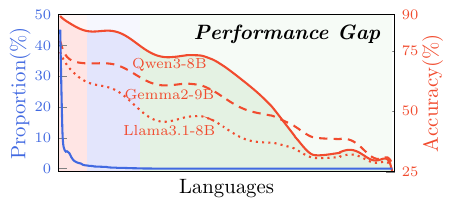}
	% \caption{subcaption1}
	% \label{fig:xxx}
\end{subfigure}
\vskip -10pt
\begin{subfigure}[]{\linewidth}
	\centering
	\includegraphics[width=\textwidth]{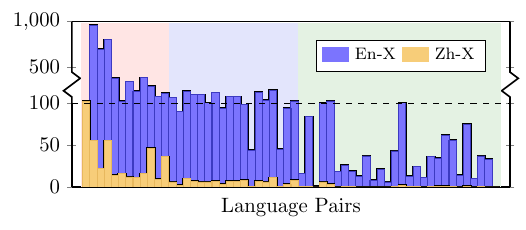}
	% \caption{subcaption1}
	% \label{fig:xxx}
\end{subfigure}

\caption{
\textbf{Top:} Performance of base LLMs (orange) on the Belebele benchmark across 108 languages, plotted against their data ratios in the CulturaX (blue). 
\textbf{Bottom:} Bilingual data volume (million sentence pairs) from the OPUS corpus for 60 languages in our study, covering English-centric (blue) and Chinese-centric (orange) directions. Languages are grouped into \colorbox{background_color1}{high-}, \colorbox{background_color2}{medium-}, and \colorbox{background_color3}{low-}resource tiers. 
}

\label{fig:lang_disparity}
\end{figure}

%% file: sections/3_method.tex
\section{The Pitfall of Directional Degeneration}
\label{sec:pitfall}

\begin{figure*}[t] 
    \centering 
    \includegraphics[width=\columnwidth]{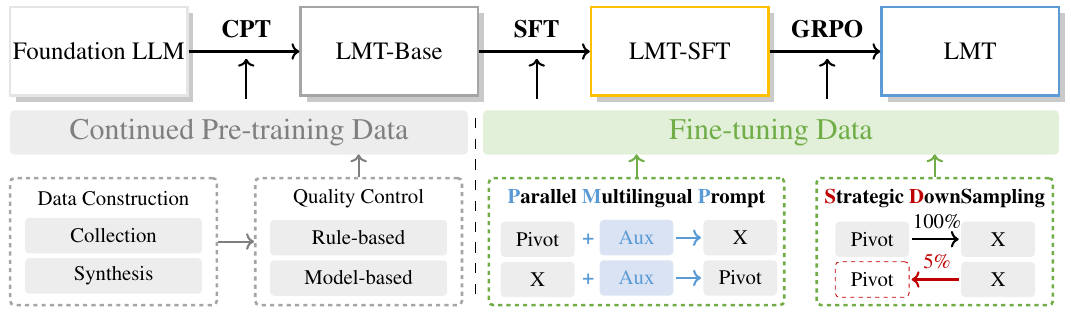} 
    \caption{Overview of the LMT framework. The pipeline consists of three stages: Continued Pre-training (CPT) on mixed monolingual and bilingual corpora, Supervised Fine-Tuning (SFT) enhanced by Parallel Multilingual Prompting (PMP) and Strategic Downsampling (SD), and Group Relative Policy Optimization (GRPO) to further refine model outputs. }
    \label{fig:framework}
\end{figure*}

\paragraph{SFT Data}
A common takeaway in instruction tuning is that high-quality, diverse supervision matters more than sheer scale \cite{DBLP:conf/nips/ZhouLX0SMMEYYZG23}.
However, for many low-resource languages, such high-quality parallel data is extremely scarce, often limited to a few multi-way benchmarks. 
Consequently, broad-coverage corpora like the FLORES-200 \cite{DBLP:journals/corr/abs-2207-04672} and NTREX-128 \cite{federmann-etal-2022-ntrex} have become indispensable, as they represent the few reliable sources of clean, human-annotated translations for these long-tail directions.
Following this widely used recipe, we take Flores-200 Devset and NTREX-128 as the primary backbone to ensure broad linguistic coverage, augment them with SMol \cite{caswell2025smol} for additional underrepresented pairs, and further add WMT14--23 and IWSLT17--24 test sets to increase style and domain diversity. 
In total, the resulting SFT dataset contains $\approx$567K high-quality parallel pairs, allocating 3K--20K examples per direction across 117 English- and Chinese-centric directions spanning 60 languages.

\paragraph{The Phenomenon and Hypothesis}
We begin by applying standard SFT to the Qwen3-4B-Base \citep{DBLP:journals/corr/abs-2505-09388} model, utilizing parallel pairs in both directions (En/Zh$\leftrightarrow$X) as per common practice.
While this standard approach yielded expected improvements in the En/Zh$\to$X directions, it unexpectedly resulted in a significant performance drop in the reverse X$\to$En/Zh directions.
Qualitative analysis shows that the model falls into a state of fluent hallucination, producing grammatically correct but factually unfaithful outputs.
\Cref{tab:error_case} shows a representative case.
We term this phenomenon \textbf{\emph{Directional Degeneration}}.
We hypothesize that this pathology stems from a \textbf{\emph{Shallow Mapping Trap}}.
The symmetric usage of multi-way data inherently creates excessive many-to-one mappings, where a single English/Chinese target is paired with dozens of distinct source languages.
This structural imbalance incentivizes the model to learn a shortcut: bypassing source semantics to simply overfit the high-frequency target patterns, thereby sacrificing faithfulness.

\paragraph{Experimental Design}
To test this hypothesis and rule out model-specific factors, we design a systematic suite of experiments along three axes: \textbf{data usage}, \textbf{model}, and \textbf{multilingual scale}. 
On the data-usage axis, we (i) \emph{Break Symmetry} by replacing the reverse X$\to$En/Zh portion of the multi-way SFT data with a completely disjoint subset sampled from our bilingual CPT corpus (symmetry-breaking replacement), and (ii) perform \emph{Gradual Symmetry Injection} by training on the original multi-way data while increasing the reverse retention rate $p$ from $0\%$ to $100\%$ (fully symmetric). 
On the model axis, we repeat this protocol on Qwen3 models ranging from 0.6B to 8B parameters, and further on Llama-3.1-8B \citep{DBLP:journals/corr/abs-2407-21783} and Gemma-2-9B \citep{DBLP:journals/corr/abs-2408-00118}, to verify whether the phenomenon persists across different model sizes and families.
On the multilingual axis, we vary the number of languages involved in SFT from 10 to 50 (using Qwen3-4B-Base) to assess how the density of many-to-one mappings influences the severity of degeneration.

\begin{figure*}[t]
\centering
\includegraphics[width=\linewidth]{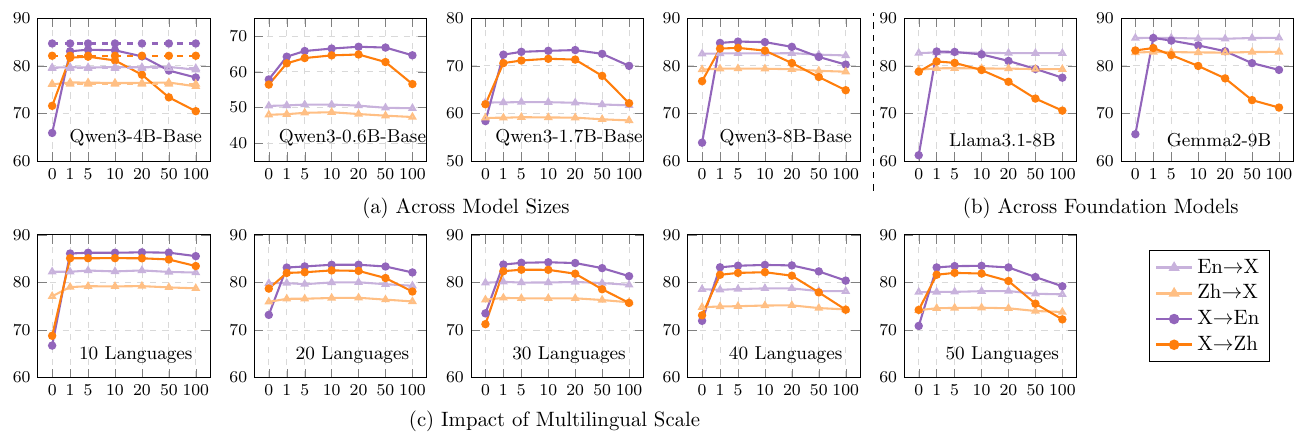}
\caption{Comprehensive analysis of Directional Degeneration across different dimensions. The plots illustrate the COMET performance trend as a function of the strategic downsampling proportion ($p$). Dashed lines in Qwen3-4B-Base represent the use of disjoint data for the X$\to$En/Zh directions.}
\label{fig:degeneration_analysis}
\end{figure*}

\paragraph{Results}
The results visualized in \cref{fig:degeneration_analysis} provide converging evidence for our hypothesis. 
First, the sharp contrast between the symmetry-breaking replacement setting (dashed lines) and fully symmetric multi-way reuse (solid curves at $100\%$) in Qwen3-4B-Base shows that the collapse is driven by how the multi-way data are reused, rather than by the intrinsic difficulty of X$\to$En/Zh directions. 
Furthermore, the gradual injection curves reveal an ``inverted-V'' trajectory: performance peaks rapidly at a low retention rate ($p \approx 5\%$) but suffers a severe decline as $p$ increases toward $100\%$, confirming that excessive target repetition triggers the degeneration.
Second, the same asymmetric pattern consistently appears across all evaluated model sizes and backbone families, indicating that directional degeneration is a general failure mode of large-scale multilingual SFT.
Finally, the degradation becomes more pronounced as we scale the number of languages, supporting the interpretation that stronger many-to-one target repetition at larger multilingual scales amplifies shortcut learning and undermines faithfulness.

\subsection{Mitigation via Strategic Downsampling}
\label{sec:sd}
While replacing reverse-direction data with disjoint data proves effective in our analysis, relying on external data sources for SFT is not always viable in practice. 
We therefore seek a self-contained solution that resolves the degeneration solely within the existing multi-way SFT corpus.
The \textit{Gradual Symmetry Injection} experiment suggests a simple yet effective mitigation strategy.
As shown by the solid curves in \cref{fig:degeneration_analysis}: while full reuse ($p{=}100\%$) leads to a clear collapse, retaining a small fraction of reverse examples is sufficient to maintain alignment without triggering the collapse.
Motivated by this, we propose Strategic Downsampling (SD): during SFT, we retain all En/Zh$\to$X data, and for the multi-way portion of the corpus we independently subsample each X$\to$En/Zh instance with probability $p$, using $p{=}5\%$ as the default setting in our models.

It is worth noting that recent work \cite{DBLP:journals/corr/abs-2502-11223} also reports this asymmetric degradation, attributes it to the curse of multilinguality, and addresses it with model-level interventions such as direction-aware training and group-wise model merging. 
Compared with this line of work, our analysis supports a more specific data-usage explanation, showing that symmetric reuse of multi-way corpora induces the degradation. 
We further find that a simple data-level strategy is sufficient to prevent it in practice.

\section{Parallel Multilingual Prompting}
Recent studies have highlighted the utility of multi-way parallel data in enhancing multilingual LLMs. 
For instance, \citet{DBLP:journals/corr/abs-2505-14045} demonstrate that utilizing multi-way corpora during CPT effectively promotes cross-lingual alignment and transfer. 
Parallelly, investigations into inference-time strategies \citep{DBLP:conf/emnlp/MuFCWLWXSLZZ24,DBLP:journals/corr/abs-2509-19770} indicate that incorporating auxiliary translations into the prompt improves performance, with \citet{DBLP:conf/emnlp/MuFCWLWXSLZZ24} further identifying specific neuronal mechanisms that support such parallel processing. 
Despite these advancements in pre-training and inference, the integration of such multi-source signals directly into the SFT for large-scale machine translation remains unexplored. 
To address this gap, we introduce Parallel Multilingual Prompting (PMP), a training strategy that incorporates auxiliary parallel context into the input prompts during SFT. 
By explicitly conditioning on both the source and an auxiliary anchor, PMP enables more direct cross-lingual transfer. 

\definecolor{background_cpt}{RGB}{237,237,237}  
\definecolor{background_sft}{RGB}{225,213,231}  
\setlength{\fboxsep}{2pt} 
\begin{figure}
    \centering
    \includegraphics[width=1.0\linewidth]{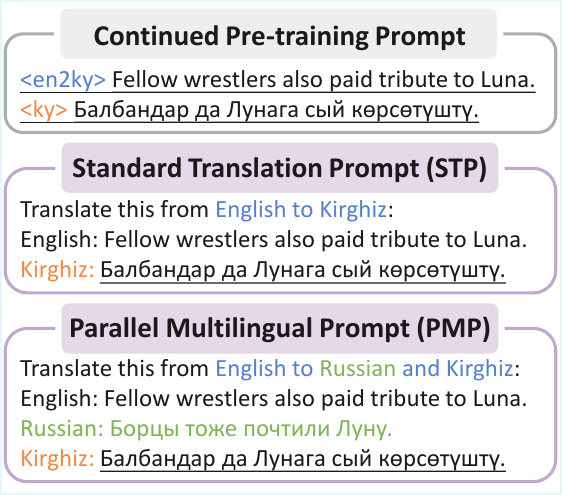}
    \caption{Examples of the three prompt formats for the \colorbox{background_cpt}{CPT} and \colorbox{background_sft}{SFT} stages of LMT adaptation. The \underline{underlined} text indicates the part used for loss computation during training.}
    \label{fig:prompt}
\end{figure}

\paragraph{Formulation}
Let $S$ be a source sentence in language $L_S$ and $T$ the target in $L_T$.
Under a standard translation prompt (STP), the model is trained to model $P_{\theta}(T \mid S;\,\tau_{L_S\!\to L_T})$, where $\tau_{L_S\!\to L_T}$ denotes the translation direction.
PMP augments the input with an auxiliary sentence $A$ in an auxiliary language $L_A$, which is itself a translation of $S$, and instead models $P_{\theta}(T \mid S, A;\,\tau_{L_S\!\to L_A\!\to L_T})$, so that $(S,A)$ provide two parallel views of the same semantic content.
Here, the auxiliary context $A$ acts as a semantic anchor that aids the model in interpreting the source $S$ and guiding the generation of the target $T$.
\cref{fig:prompt} compares these two prompting strategies.

\paragraph{Auxiliary Language Selection}
To select the most effective auxiliary language, our strategy considers two primary factors: linguistic affinity and model proficiency. 
Intuitively, an anchor is most beneficial when it shares close linguistic ties with the specific language being supported while being well-mastered by the model. 
Following this rationale: For En $\leftrightarrow$ X, we look for a high-proficiency neighbor: we select a language that shares close typological ties with X (e.g., via script, phylogeny, or contact) and on which the model demonstrates robust competence.
For Zh$\leftrightarrow$X, we use English ($L_A{=}\mathrm{En}$), which the model typically handles most reliably and thus provides a stable semantic anchor. 
This choice also facilitates test-time PMP, since self-generating an English anchor is typically easier and more robust than generating anchors in other languages.
Detailed mapping information are provided in Appendix~\cref{tab:auxiliary}.

\paragraph{Training and Inference}
We integrate PMP into SFT by probabilistically mixing PMP samples with standard STP samples.
This explicitly trains the model to exploit auxiliary parallel context while preserving performance under STP.
At inference, the model translates using the standard STP prompt by default.
When an auxiliary sentence $A$ is provided (e.g., from a high-quality external MT system or self-generated), we can use the PMP prompt to activate the learned behavior, which can yield additional improvements in translation quality.

\section{The LMT Framework}
LMT is a suite of Chinese–English-centric MMT models scaled across four distinct sizes (0.6B, 1.7B, 4B, and 8B).
The suite covers 60 languages spanning diverse families and scripts, supporting a total of 234 translation directions (including English $\leftrightarrow$ 59 languages and Chinese $\leftrightarrow$ 58 languages).
The complete language list is available in the Appendix.
Guided by our preliminary comparison in \cref{fig:lang_disparity} (top), which shows that Qwen3 offers a more comprehensive multilingual capability than contemporary open models with similar scale, we adopt Qwen3 as the backbone for all LMT variants.

\subsection{Adaptation Pipeline}
We adapt the base models to the MMT task through a three-stage pipeline, applied consistently across all four model scales to produce the LMT-60 suite:
\begin{itemize}[leftmargin=*] 
    \item \textbf{Stage 1: Continued Pre-training (CPT).} We first train the Qwen3 base models on a large-scale mixture of monolingual and parallel corpora to strengthen the model's broad multilingual and translation knowledge. 
    \item \textbf{Stage 2: Supervised Fine-tuning (SFT).} We instruction-tune the model on the curated STF data (Section~\ref{sec:pitfall}). During SFT, we apply SD to the multi-way portion to prevent directional degeneration, and incorporate PMP to train the model to leverage auxiliary parallel context.
    \item \textbf{Stage 3: Preference Optimization (PO).} We further refine the model with GRPO, reusing the SFT prompts to sample candidate translations and scoring them with COMET-22 as a reference-based reward. This setup improves quality without introducing extra preference data.
\end{itemize}

\setlength{\fboxsep}{1pt} 
\renewcommand{\arraystretch}{1.2}
\begin{table*}[ht]
  \centering
  \begin{adjustbox}{max width=\textwidth}
  \setlength{\tabcolsep}{3pt}
  \begin{tabular}{lcccccccccccc}
    \toprule
     \multirow{2.6}{*}{\textbf{Model}} &\multicolumn{4}{c}{\textbf{High Resource }} &\multicolumn{4}{c}{\textbf{Medium Resource }} &\multicolumn{4}{c}{\textbf{Low Resource }} \\ 
     \cmidrule(lr){2-5} \cmidrule(lr){6-9} \cmidrule(lr){10-13}
    &\textbf{En$\to$X} &\textbf{Zh$\to$X} &\textbf{X$\to$En} &\textbf{X$\to$Zh}
    &\textbf{En$\to$X} &\textbf{Zh$\to$X} &\textbf{X$\to$En} &\textbf{X$\to$Zh}
    &\textbf{En$\to$X} &\textbf{Zh$\to$X} &\textbf{X$\to$En} &\textbf{X$\to$Zh} \\
    \specialrule{0.08em}{0pt}{0pt}
    Qwen3-4B-Base & 84.63 & 80.77 & 85.87 & 85.44 & 79.68 & 76.21 & 86.09 & 84.55 & 56.81 & 53.33 & 75.36 & 75.35 \\
    \hdashline
    \hspace{0.2em} SFT &
      87.72$_{\textcolor{mygain}{+3.09}}$ & 85.11$_{\textcolor{mygain}{+4.34}}$ & 83.82$_{\textcolor{myloss}{-2.05}}$ & 73.60$_{\textcolor{myloss}{-11.84}}$ &
      86.71$_{\textcolor{mygain}{+7.03}}$ & 83.58$_{\textcolor{mygain}{+7.37}}$ & 80.00$_{\textcolor{myloss}{-6.09}}$ & 72.18$_{\textcolor{myloss}{-12.37}}$ &
      77.51$_{\textcolor{mygain}{+20.70}}$ & 73.68$_{\textcolor{mygain}{+20.35}}$ & 73.78$_{\textcolor{myloss}{-1.58}}$&  67.94$_{\textcolor{myloss}{-7.41}}$ \\
    \hspace{0.2em} + SD &
      87.80$_{+0.08}$ & 85.32$_{+0.21}$ & 87.49$_{\textcolor{mygain}{+3.67}}$ & 86.55$_{\textcolor{mygain}{+12.95}}$ &
      86.72$_{+0.01}$ & 83.67$_{+0.09}$ & 87.67$_{\textcolor{mygain}{+7.67}}$ & 85.87$_{\textcolor{mygain}{+13.69}}$ &
      78.68$_{\textcolor{mygain}{+1.17}}$ & 75.15$_{\textcolor{mygain}{+1.47}}$ & 80.72$_{\textcolor{mygain}{+6.94}}$ & 79.13$_{\textcolor{mygain}{+11.19}}$ \\
    \hspace{0.2em} + CPT &
      89.03$_{\textcolor{mygain}{+1.23}}$ & 86.72$_{\textcolor{mygain}{+1.40}}$ & 87.97$_{+0.48}$ & 87.39$_{+0.84}$ &
      89.77$_{\textcolor{mygain}{+3.05}}$ & 86.96$_{\textcolor{mygain}{+3.29}}$ & 88.56$_{+0.89}$ & 87.06$_{\textcolor{mygain}{+1.19}}$ &
      87.14$_{\textcolor{mygain}{+8.46}}$ & 84.17$_{\textcolor{mygain}{+9.02}}$ & 86.02$_{\textcolor{mygain}{+5.30}}$ & 84.74$_{\textcolor{mygain}{+5.61}}$ \\
    \hspace{0.2em} + PMP &
      88.98$_{-0.05}$ & 86.74$_{+0.02}$ & 88.00$_{+0.03}$ & 87.53$_{+0.14}$ &
      89.73$_{-0.04}$ & 86.92$_{-0.04}$ & 88.62$_{+0.06}$ & 87.20$_{+0.14}$ &
      87.06$_{-0.08}$ & 84.08$_{-0.09}$ & 86.07$_{+0.05}$ & 84.90$_{+0.16}$ \\
    \rowcolor{gray!20} \hspace{0.2em} + GRPO &
      \textbf{89.43}$_{+0.45}$ & \textbf{87.20}$_{+0.46}$ & \textbf{88.46}$_{+0.46}$ & \textbf{88.19}$_{+0.66}$ &
      \textbf{90.23}$_{+0.50}$ & \textbf{87.52}$_{+0.60}$ & \textbf{89.10}$_{+0.48}$ & \textbf{87.97}$_{+0.77}$ &
      \textbf{87.85}$_{+0.79}$ & \textbf{84.92}$_{+0.84}$ & \textbf{86.60}$_{+0.53}$ & \textbf{85.81}$_{+0.91}$ \\
    \specialrule{0.05em}{0pt}{0pt}
    %%%%%%%%%%%%%%%%%%%%%%%%%%%%%%%%%%%
    Qwen3-8B-Base & 87.02 & 83.55 & 86.65 & 86.05 & 84.69 & 81.98 & 87.29 & 85.60 & 66.43 & 63.32 & 78.50 & 78.59 \\
    \hdashline
    \hspace{0.2em} SFT &
      88.53$_{\textcolor{mygain}{+1.51}}$ & 85.96$_{\textcolor{mygain}{+2.41}}$ & 84.86$_{\textcolor{myloss}{-1.79}}$ & 77.83$_{\textcolor{myloss}{-8.22}}$ &
      88.21$_{\textcolor{mygain}{+3.52}}$ & 85.22$_{\textcolor{mygain}{+3.24}}$ & 82.27$_{\textcolor{myloss}{-5.02}}$ & 76.55$_{\textcolor{myloss}{-9.05}}$ &
      80.72$_{\textcolor{mygain}{+14.29}}$ & 76.95$_{\textcolor{mygain}{+13.63}}$ & 77.61$_{\textcolor{myloss}{-0.89}}$ & 73.08$_{\textcolor{myloss}{-5.51}}$\\
    \hspace{0.2em} + SD &
      88.57$_{+0.04}$ & 86.13$_{+0.17}$ & 87.69$_{\textcolor{mygain}{+2.83}}$ & 86.84$_{\textcolor{mygain}{+9.01}}$ &
      88.28$_{+0.07}$ & 85.37$_{+0.15}$ & 87.99$_{\textcolor{mygain}{+5.72}}$ & 86.33$_{\textcolor{mygain}{+9.78}}$ &
      82.49$_{\textcolor{mygain}{+1.77}}$ & 79.12$_{\textcolor{mygain}{+2.17}}$ & 82.83$_{\textcolor{mygain}{+5.22}}$ & 81.52$_{\textcolor{mygain}{+8.44}}$ \\
    \hspace{0.2em} + CPT &
      89.31$_{+0.74}$ & 87.07$_{+0.94}$ & 88.02$_{+0.33}$ & 87.46$_{+0.62}$ &
      90.06$_{\textcolor{mygain}{+1.78}}$ & 87.35$_{\textcolor{mygain}{+1.98}}$ & 88.57$_{+0.58}$ & 87.17$_{+0.84}$ &
      87.42$_{\textcolor{mygain}{+4.93}}$ & 84.51$_{\textcolor{mygain}{+5.39}}$ & 86.32$_{\textcolor{mygain}{+3.49}}$ & 85.18$_{\textcolor{mygain}{+3.66}}$ \\
    \hspace{0.2em} + PMP &
      89.29$_{-0.02}$ & 87.10$_{+0.03}$ & 88.06$_{+0.04}$ & 87.60$_{+0.14}$ &
      90.06$_{+0.00}$ & 87.28$_{-0.07}$ & 88.63$_{+0.06}$ & 87.39$_{+0.22}$ &
      87.38$_{-0.04}$ & 84.50$_{-0.01}$ & 86.41$_{+0.09}$ & 85.41$_{+0.23}$ \\
    \rowcolor{gray!20} \hspace{0.2em} + GRPO &
      \textbf{89.60}$_{+0.31}$ & \textbf{87.41}$_{+0.31}$ & \textbf{88.50}$_{+0.44}$ & \textbf{88.22}$_{+0.62}$ &
      \textbf{90.39}$_{+0.33}$ & \textbf{87.70}$_{+0.42}$ & \textbf{89.10}$_{+0.47}$ & \textbf{87.95}$_{+0.56}$ &
      \textbf{87.93}$_{+0.55}$ & \textbf{85.04}$_{+0.54}$ & \textbf{86.91}$_{+0.50}$ & \textbf{86.08}$_{+0.67}$ \\
    \specialrule{0.08em}{0pt}{0pt}
  \end{tabular}
  \end{adjustbox}
  \caption{COMET-22 scores of 4B and 8B models as we progressively enable components of the LMT training pipeline: supervised fine-tuning (SFT), Strategic Downsampling (SD), continued pre-training (CPT), Parallel Multilingual Prompting (PMP), and preference optimization (PO). 
\textbf{Bold} numbers indicate the best score within each backbone block. Subscripts denote the score difference compared to the previous row. We use \textcolor{myloss}{red} to mark entries affected by \emph{directional degeneration}, and \textcolor{mygain}{green} to highlight substantial improvements (>1.0).}
  \label{tab:main_result}
\end{table*}
\renewcommand{\arraystretch}{1}

\subsection{CPT Data Curation}
To support effective CPT, we construct a multi-stage data pipeline comprising large-scale collection, pseudo-parallel synthesis, and systematic filtering. 
For monolingual data, we aggregate text for the 60 target languages from a broad range of public, curated multilingual sources, and apply standard cleaning and de-duplication.
For parallel data, we start from OPUS and examine the available parallel volume for our 117 directions.
As illustrated in \cref{fig:lang_disparity} (bottom), we observe a significant imbalance: while English-centric directions are relatively well-covered, Chinese-centric directions suffer from severe data scarcity.
To bridge this gap, we extensively augment the authentic parallel pool with synthetic data generated by high-performing MT systems. 
Following a systematic quality filtering process, we obtain a total of approximately 2.1B English-centric and 2.9B Chinese-centric sentence pairs across 117 language pairs, which lay a solid data foundation for subsequent adaptation. 
Detailed corpus composition, quality estimation, and statistics are provided in Appendix~\ref{sec:cpt_data}.

%% file: sections/4_results.tex
\section{Results and Analyses}

\subsection{Setup}
\paragraph{Training Details}
For CPT, we use 90B tokens balanced at a 1:1:1 ratio across monolingual, Chinese-centric bilingual, and English-centric bilingual data.
The bilingual samples are used equally in both directions (En/Zh$\to$X and X$\to$En/Zh; 50/50), and adopt an \textbf{Informative Formatting} with explicit direction tags and a target-language separator (as shown in \cref{fig:prompt}), which we find performs better slightly than naïve newline source–target concatenation \citep{DBLP:conf/naacl/GuoYLWSC24,DBLP:conf/wmt/IyerMSCHB24}.
During SFT, for forward directions (En/Zh$\to$X), STP and PMP each account for 50\%.
For reverse directions (X $\to$ En/Zh), we apply strategic downsampling with a total retention of 5\%, split evenly between formats (STP 2.5\%, PMP 2.5\%).
Finally, we perform preference optimization with GRPO by reusing SFT prompts to generate rollouts and scoring them with COMET-22 ~\cite{DBLP:conf/wmt/ReiSAZFGLCM22} as a reference-based reward.
We train four model sizes (0.6B/1.7B/4B/8B) on 16 NVIDIA H200 GPUs, and the detailed hyperparameters are provided in the Appendix \cref{tab:hyperparameters}.

\paragraph{Evaluation Data and Metrics}  
We evaluate on FLORES-200 Devtest \citep{DBLP:journals/corr/abs-2207-04672}.
To address the lack of a Mongolian (traditional script) testset, we translated the Chinese side of FLORES into Mongolian with native annotators \footnote{We release this dataset to fill a gap in MT Benchmark.}. 
We adopt COMET-22 as our primary evaluation metric, and report SacreBLEU~\citep{post-2018-call} in the Appendix.
For brevity, we present LMT-60-4/8B in the main text, with full results for all four sizes  provided in the Appendix.
In addition, we also provide a comparison against the WMT24++  \citep{deutsch2025wmt24expandinglanguagecoverage} in the Appendix.

\subsection{Main Results}
\label{sec:main_results}

\cref{tab:main_result} summarizes the evolution of translation quality as we progressively integrate components of the LMT training pipeline. 
We report COMET-22 scores averaged over all 60 languages, categorized by resource tiers and translation direction.
Starting from the 3-shot base models, SFT improves En/Zh$\to$X on both 4B and 8B settings, but all X$\to$En/Zh drop significantly below the corresponding base scores, as indicated by the \textcolor{myloss}{red subscripts}.
This empirically confirms that SFT under symmetric multi-way reuse can trigger severe directional degeneration in reverse directions.
Integrating SD effectively reverses this effect: X$\to$En/Zh directions typically recover by approximately 2–13 COMET points relative to the SFT baseline, surpassing base performance while maintaining strong results on En/Zh$\to$X.

Compared to the SD+SFT baseline, CPT yields substantial improvements across the board. 
It contributes 1–3 COMET points in high- and medium-resource scenarios and a remarkable 5–9 points in low-resource directions, underscoring its critical role in strengthening the model’s broad translation ability. \footnote{Detailed analyses are provided in Appendix~\ref{sec:appendix_cpt_analysis}.}
The introduction of PMP yields gains on X$\to$En/Zh directions, likely by enriching source-side diversity to alleviate the directional degeneration.
Finally, GRPO reuses the same SFT examples but still brings a further gain of about 0.3–0.8 points across resource tiers. 
This indicates that preference optimization can extract additional benefit from the supervised data by exploring alternative generations and reinforcing better candidates, even when no new training examples are introduced.

\definecolor{background_trans}{RGB}{227,229,252}  % dedicated MMT
\definecolor{background_general}{RGB}{228,242,227}  % general LLM

\setlength{\fboxsep}{1pt} 
\renewcommand{\arraystretch}{1.2}
\begin{table}[t]
  \centering
  \begin{adjustbox}{max width=\columnwidth}
  \setlength{\tabcolsep}{3pt}
  \begin{tabular}{clccccc}
    \toprule
     \textbf{\#Langs} & \textbf{Model} & \textbf{En$\to$X} & \textbf{X$\to$En} & \textbf{Zh$\to$X} & \textbf{X$\to$Zh} & \textbf{Avg.} \\
    \specialrule{0.05em}{0pt}{0pt}
    %%%%%%%%%%%%%%%%%%%%%%%%%%%%%%%%%%%%%%%%%%%
    \cellcolor{background_trans}
    & TowerInstruct-13B & 88.91 & 88.51 & 86.29 & 86.81 & 87.63 \\
    \cellcolor{background_trans}
    10 & LMT-60-4B & \underline{89.29} & \underline{88.58} & \underline{87.09} & \textbf{88.40} & \underline{88.34} \\
    \cellcolor{background_trans}
    & LMT-60-8B & \textbf{89.42} & \textbf{88.59} & \textbf{87.30} & \textbf{88.40} & \textbf{88.43} \\
    \specialrule{0.05em}{0pt}{0pt}
    %%%%%%%%%%%%%%%%%%%%%%%%%%%%%%%%%%%%%%%%%%%
    \cellcolor{background_general}
    & Aya-expanse-8B & 88.66 & 88.31 & 86.26 & 86.20 & 87.36  \\
    \cellcolor{background_general}
    23 & LMT-60-4B & \underline{89.44} & \underline{88.62} & \underline{87.00} & \textbf{87.99} & \underline{88.26} \\
    \cellcolor{background_general}
    & LMT-60-8B & \textbf{89.63} & \textbf{88.65} & \textbf{87.18} & \underline{87.98} & \textbf{88.36} \\
    \specialrule{0.05em}{0pt}{0pt}
    %%%%%%%%%%%%%%%%%%%%%%%%%%%%%%%%%%%%%%%%%%%
    \cellcolor{background_trans}
    & Seed-X-PPO-7B & \textbf{90.78} & \textbf{89.05} & \textbf{88.48} & 87.96 & \textbf{89.07} \\
    \cellcolor{background_trans}
    27 & LMT-60-4B & 90.35 & 88.87 & 88.02 & \textbf{88.19} & 88.86 \\
    \cellcolor{background_trans}
    & LMT-60-8B & \underline{90.49} & \underline{88.88} & \underline{88.23} & \underline{88.16} & \underline{88.94} \\
    \specialrule{0.05em}{0pt}{0pt}
    %%%%%%%%%%%%%%%%%%%%%%%%%%%%%%%%%%%%%%%%%%%
    \cellcolor{background_trans}
    & GemmaX2-28-9B & 88.39 & \textbf{88.95} & 85.56 & 87.37 & 87.57 \\
    \cellcolor{background_trans}
    28 & LMT-60-4B & \underline{88.72} & 88.52 & \underline{85.97} & \underline{87.71} & \underline{87.73} \\
    \cellcolor{background_trans}
    & LMT-60-8B & \textbf{88.83} & \underline{88.62} & \textbf{86.12} & \textbf{87.76} & \textbf{87.83} \\
    \specialrule{0.05em}{0pt}{0pt}
    %%%%%%%%%%%%%%%%%%%%%%%%%%%%%%%%%%%%%%%%%%%
    \cellcolor{background_trans}
    & Hunyuan-MT-7B & 86.78 & 86.42 & 83.88 & 85.74 & 85.71 \\
    \cellcolor{background_trans}
    35 & LMT-60-4B & \underline{88.72} & \underline{88.00} & \underline{86.03} & \underline{87.26} & \underline{87.50} \\
    \cellcolor{background_trans}
    & LMT-60-8B & \textbf{88.84} & \textbf{88.12} & \textbf{86.18} & \textbf{87.36} & \textbf{87.63} \\
    \specialrule{0.05em}{0pt}{0pt}
    %%%%%%%%%%%%%%%%%%%%%%%%%%%%%%%%%%%%%%%%%%%
    \cellcolor{background_trans}
    & X-ALMA-13B & 89.17 & \underline{88.68} & - & - & 88.92 \\
    \cellcolor{background_trans}
    40 & LMT-60-4B & \underline{89.24} & 88.67 & - & - & \underline{88.96} \\
    \cellcolor{background_trans}
    & LMT-60-8B & \textbf{89.38} & \textbf{88.73} & - & - & \textbf{89.06} \\
    \specialrule{0.05em}{0pt}{0pt}
    %%%%%%%%%%%%%%%%%%%%%%%%%%%%%%%%%%%%%%%%%%%
    \cellcolor{background_general}
    & Aya-101-13B & 84.87 & 86.45 & 81.53 & 82.54 & 83.85 \\
    \cellcolor{background_general}
    54 & LMT-60-4B & \underline{88.61} & \underline{88.23} & \underline{85.67} & \underline{87.15} & \underline{87.42} \\
    \cellcolor{background_general}
    & LMT-60-8B & \textbf{88.75} & \textbf{88.36} & \textbf{85.84} & \textbf{87.24} & \textbf{87.55} \\
    \specialrule{0.05em}{0pt}{0pt}
    %%%%%%%%%%%%%%%%%%%%%%%%%%%%%%%%%%%%%%%%%%%
    \cellcolor{background_general}
    & LLaMAX3-Alpaca & 81.29 & 86.27 & 77.24 & 81.02 & 81.45 \\
    \cellcolor{background_general}
    55 & LMT-60-4B & \underline{88.66} & \underline{88.20} & \underline{85.74} & \underline{87.16} & \underline{87.44} \\
    \cellcolor{background_general}
    & LMT-60-8B & \textbf{88.78} & \textbf{88.32} & \textbf{85.91} & \textbf{87.23} & \textbf{87.56} \\
    \specialrule{0.05em}{0pt}{0pt}
    %%%%%%%%%%%%%%%%%%%%%%%%%%%%%%%%%%%%%%%%%%%
    \cellcolor{background_trans}
    & NLLB-54B & 86.89 & 87.72 & 84.06 & 80.50 & 84.79 \\
    \cellcolor{background_trans}
    59 & LMT-60-4B & \underline{88.78} & \underline{87.93} & \underline{86.00} & \underline{87.00} & \underline{87.43} \\
    \cellcolor{background_trans}
    & LMT-60-8B & \textbf{88.91} & \textbf{88.07} & \textbf{86.16} & \textbf{87.11} & \textbf{87.56} \\
    \specialrule{0.08em}{0pt}{0pt}
  \end{tabular}
  \end{adjustbox}
\caption{
COMET-22 scores of our LMT models compared with a range of \colorbox{background_general}{general-purpose} multilingual LLMs and \colorbox{background_trans}{dedicated MMT} models, averaged over all overlapping languages for each system in four directions. ``\#Langs'' denotes the number of languages shared between the baseline and LMT. \textbf{Bold} numbers indicate the best score in each comparison group, and \underline{underlined} numbers the second best. The symbol ``-'' indicates directions not supported by the baseline model.
}
\label{tab:main_result_compact}
\end{table}

\renewcommand{\arraystretch}{1.0}

\subsection{Comparison with Existing MMT Systems}
\label{sec:comparison}

To benchmark LMT against the current SOTA, \cref{tab:main_result2} presents a comprehensive comparison with a diverse range of systems, categorized into two groups: (1) \textit{General-purpose Multilingual LLMs} capable of instruction-following translation, including Aya-Expanse-8B \citep{dang2024ayaexpansecombiningresearch}, Aya-101-13B \citep{2024aya}, and LLaMAX3-8B-Alpaca \citep{lu-etal-2024-llamax}; (2) \textit{Dedicated MMT Models}, including TowerInstruct-13B \citep{DBLP:journals/corr/abs-2402-17733}, GemmaX2-28-9B \citep{DBLP:conf/naacl/CuiGLLW25}, X-ALMA-13B \citep{DBLP:conf/iclr/XuMKHEK25}, Hunyuan-MT-7B \citep{zheng2025hunyuanmt}, Seed-X-PPO-7B \citep{DBLP:journals/corr/abs-2507-13618}, and NLLB-54B \citep{DBLP:journals/corr/abs-2207-04672}. To ensure a fair evaluation, we calculate metrics solely on the intersection of languages supported by each baseline, reporting the averaged COMET-22 scores over overlapping directions. 
We report a resource-tier breakdown on FLORES and additional results in Appendix \ref{sec:tier_wmt24}.

The results demonstrate that LMT models deliver robust and consistent performance across all comparison groups.
Against general-purpose multilingual LLMs, LMT achieves substantial improvements, surpassing models like Aya-101-13B and LLaMAX3-Alpaca by a margin of approximately 3–6 COMET points on average. 
Compared to dedicated MMT systems, LMT remains highly competitive. 
It outperforms strong baselines such as NLLB-54B and GemmaX2-28-9B, and performs on par with top-tier systems like Seed-X-PPO-7B. 
Most notably, LMT-60-4B demonstrates exceptional parameter efficiency, matching or exceeding the performance of significantly larger models (e.g., the Aya-101-13B and NLLB-54B). 

Overall, these results position LMT as a competitive MMT baseline: it covers a broader set of languages than most existing LLM-based MT systems, and achieves comparable or better quality than much larger baselines on the overlapping directions.

\begin{figure}
    \centering
    \includegraphics[width=\linewidth]{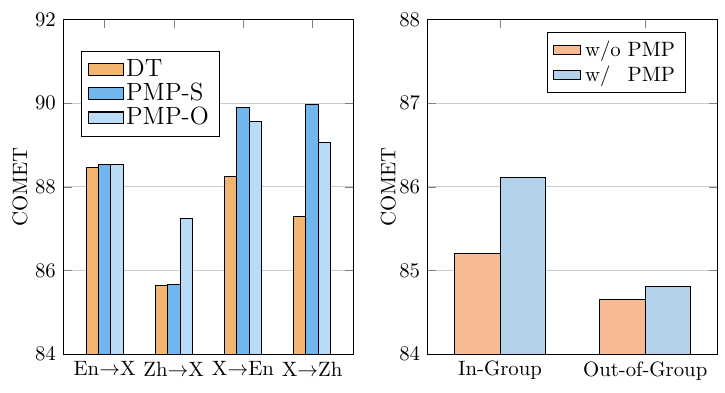}
    \caption{Analysis of the Parallel Multilingual Prompting (PMP). \textbf{Left:} Comparison of different inference-time strategies. \textbf{Right:} Comparison of zero-shot performance with and without PMP training.}
    \label{fig:pmp_analysis}
\end{figure}

\subsection{Understanding the Effect of PMP}

\paragraph{Test-time Enhancement}
We first analyze the effect of PMP at inference time on the PMP-enabled languages (see \cref{tab:auxiliary}).
We compare three modes: Direct Translation (DT), which uses the standard translation prompt (STP), PMP with a self-generated auxiliary sentence (PMP-S), and PMP with an oracle auxiliary sentence (PMP-O).
As illustrated in \cref{fig:pmp_analysis} (left), the effectiveness of PMP varies by direction.
On X$\to$En/Zh, PMP-S performs comparably to or better than PMP-O, suggesting that for high-resource targets, the model is resilient to imperfect anchors; self-generated context is sufficient to boost performance.
Conversely, on Zh$\to$X, we find that only PMP-O yields clear gains, while PMP-S does not. 
This implies that translating into X is sensitive to anchor quality: while gold-standard context helps, noisy self-generated anchors may fail to provide positive guidance.
Practically, these findings highlight PMP-S as a robust inference-time enhancement strategy for X$\to$En/Zh directions, enabling the model to self-boost performance without relying on external references.

\paragraph{PMP Improves Zero-shot Transfer}
Beyond inference benefits, we examine whether PMP training enhances the model’s cross-lingual transfer in zero-shot directions.
We compare models trained with and without PMP on two distinct subsets of directions among the participating languages (\cref{fig:pmp_analysis} (right)).
In the \emph{In-Group} setting (translations between auxiliary languages and their associated targets, A$\leftrightarrow$X), PMP training substantially boosts the average COMET-22 score from 85.20 to 86.11.
Furthermore, in the \emph{Out-of-Group} setting, where we evaluate directions between high-resource languages never explicitly used as PMP anchors (e.g., \{Es, Ja\}$\leftrightarrow$X), we still observe a positive performance uplift.
These results indicate that PMP not only strengthens the anchored language pairs but also promotes zero-shot transfer to related directions that were never explicitly used as auxiliary anchors during training.

%% file: sections/2_related_work.tex
\section{Related Work}

\paragraph{LLM Adaptation for MMT}
While dedicated encoder-decoder models like NLLB \citep{DBLP:journals/corr/abs-2207-04672} and M2M-100 \citep{DBLP:journals/jmlr/FanBSMEGBCWCGBL21} have long served as the backbone of multilingual translation, recent trends have shifted towards adapting decoder-only LLMs for this task \citep{DBLP:journals/corr/abs-2305-18098,DBLP:journals/corr/abs-2402-17733,DBLP:conf/iclr/XuMKHEK25,DBLP:conf/naacl/CuiGLLW25,luoyf2025lamate}.
This transition brings not only stronger translation quality but also advanced capabilities such as context awareness \citep{DBLP:conf/acl/WangDJLPCSW0T24,DBLP:conf/iclr/WangZLWMZZ25} and reasoning \citep{DBLP:conf/acl/WangMLZ25,DBLP:journals/corr/abs-2502-11544}.
Despite this progress, current LLM-based MMT systems still face several limitations.
First, regarding language coverage, most existing adaptations remain restricted to a narrow set of dominant languages, struggling to match the extensive linguistic breadth required for truly universal translation.
Second, regarding translation directions, the research landscape remains heavily English-centric. 
Consequently, non-English directions are largely underserved, resulting in inferior performance compared to English-centric tasks, with only a few recent works extending capabilities to Chinese-centric translation\citep{DBLP:conf/naacl/CuiGLLW25,zheng2025hunyuanmt,DBLP:journals/corr/abs-2507-13618}.
Finally, low-resource translation remains a persistent challenge for adapted LLMs. 
Under severe data scarcity and the dominance of high-resource languages in pre-training, these models often struggle to maintain robustness on resource-poor languages.

\paragraph{Utilization of Multi-way Data}
Multi-way parallel data, where aligned sentences exist across three or more languages, provides richer semantic constraints than bilingual counterparts.
In traditional NMT, this concept was explored through multi-source translation to improve disambiguation and robustness \citep{DBLP:journals/taslp/NishimuraSNN20,DBLP:conf/emnlp/XuYM0H21}.
In the era of LLMs, recent analysis has begun to reveal the parallel multilingual learning mechanisms within these models, suggesting that LLMs inherently benefit from cross-lingual alignment \citep{DBLP:conf/emnlp/MuFCWLWXSLZZ24,DBLP:journals/corr/abs-2505-14045,DBLP:journals/corr/abs-2509-19770}.
However, while these works demonstrate the value of multi-way signals, how to properly leverage them for consistent gains in large-scale multilingual machine translation, and the potential pitfalls of widely used symmetric multi-way mixtures in SFT remain under-explored.

%% file: sections/5_conclusion.tex
\section{Conclusion}
In this work, we introduce LMT, a Chinese-English-centric  MMT model, covering 60 languages across 234 translation directions and achieving competitive performance among models with similar language coverage.
Our adaptation pipeline begins with the construction of a large, curated multilingual corpus, followed by extensive continued pre-training to integrate broad translation knowledge into the model.
We then identify directional degeneration, a salient yet previously overlooked issue in SFT with multi-way data, and mitigate it by introducing a simple strategic downsampling method.
Furthermore, we propose Parallel Multilingual Prompting, a simple but effective technique to enhance cross-lingual transfer.
We release LMT models as publicly available baselines to facilitate future research on inclusive and high-quality multilingual machine translation.

\section*{Limitations}
While LMT shows promising performance, we note several limitations that also point to natural directions for future work.
First, although our evaluation covers a broad set of standard academic benchmarks and primarily relies on COMET, these settings may not fully reflect the diversity of real-world translation use cases. 
Future work could extend this evaluation to a wider range of real-world scenarios to further assess generalization and capture more nuanced aspects of translation quality. 
Second, LMT adopts a Chinese–English-centric design as a step toward moving beyond an English-only focus. 
This bi-centric setting is still a simplifying choice and may not be optimal for other regions or language communities. 
It would be useful to explore tri-centric or more general multi-centric configurations, and to study how they affect scalability, interference, and cross-lingual transfer.
Third, LMT currently supports 60 languages, which remains a small subset of global linguistic diversity. 
Expanding coverage is constrained not only by training cost but also by the availability and quality of text and parallel data for many underrepresented languages, especially those with limited written resources. 
Future work could prioritize extending support to additional languages and improving adaptation under extreme data scarcity through more effective data collection, filtering, and transfer strategies.

\section*{Acknowledgments}
This work was supported in part by the National Natural Science Foundation of China (Nos. U24A20334 and 62276056), the Yunnan Fundamental Research Projects (No.202401BC070021), the Yunnan Science and Technology Major Project (No. 202502AD080014), the Fundamental Research Funds for the Central Universities (Nos. N25BSS054 and N25BSS094), and the Program of Introducing Talents of Discipline to Universities, Plan 111 (No.B16009).
We are also grateful to the members of our team for their contributions to data collection and cleaning, system development, and experimental evaluation.

%% file: appendices/a_setup.tex
\clearpage
\section{Continued Pre-training Data Curation}
\label{sec:cpt_data}

The foundation of CPT relies on a high-quality, large-scale, and diverse corpus. 
Recognizing that constructing such a corpus is a major challenge, we designed a systematic, multi-stage data curation pipeline to ensure both linguistic breadth and quality consistency.

\subsection{CPT Monolingual Data}
To achieve comprehensive coverage and diversity, our monolingual CPT corpus aggregates several sources. 
For English and Chinese, we utilized the well-curated SlimPajama \cite{cerebras2023slimpajama} and Skywork \cite{wei2023skywork} corpora, respectively.
For the remaining languages, we collected from CulturaX \cite{nguyen-etal-2024-culturax}, OpenDataLab \cite{yu2025wanjuansiluhighqualityopensourcewebtext}, and Wikimedia, with full source details in ~\cref{tab:lang_source}.

\subsection{CPT Bilingual Data}
The foundation of the bilingual corpus is curated from OPUS \footnote{https://opus.nlpl.eu/} sub-corpora. 
To significantly scale this up, we employed pseudo-parallel synthesis using open-source models in two ways: (1) direct synthesis, creating synthetic En/Zh $\to$ X data by translating monolingual corpora, and (2) pivoted synthesis via English, leveraging typically higher-quality En$\leftrightarrow$X and En$\rightarrow$Zh models to obtain Zh$\leftrightarrow$X data.
Finally, all data was unified and subjected to a rigorous quality control pipeline. 
We used OpusFilter \citep{aulamo-etal-2020-opusfilter} for rule-based cleaning, including length-based sanity checks and misalignment removal; in particular, we filtered out sentence pairs with a maximum length ratio greater than 3.0.
For language identification, we employed the standard FastText LID model integrated in OpusFilter, with an adaptive strategy to accommodate different resource levels: for high-resource languages we enforced a confidence threshold of 0.5 to ensure data purity, while for low-resource languages (where LID probabilities are often poorly calibrated) we prioritized recall by accepting samples as long as the Top-1 predicted language matched the target tag, regardless of its probability.
We further applied CometKiwi \citep{DBLP:conf/wmt/ReiTGZFMSGACLM22} for quality-based scoring and selection.
This yields approximately 2.1B sentence pairs for English-centric and 2.9B for Chinese-centric directions, with the vast majority comprising over 10M high-quality pairs across the 117 targeted directions.

\subsection{Quality Analysis of the Bilingual Corpus}
\begin{center}
\begin{figure*}[h]
    \centering
    \begin{adjustbox}{width=1.1\textwidth, center}
    \includegraphics[width=\linewidth]{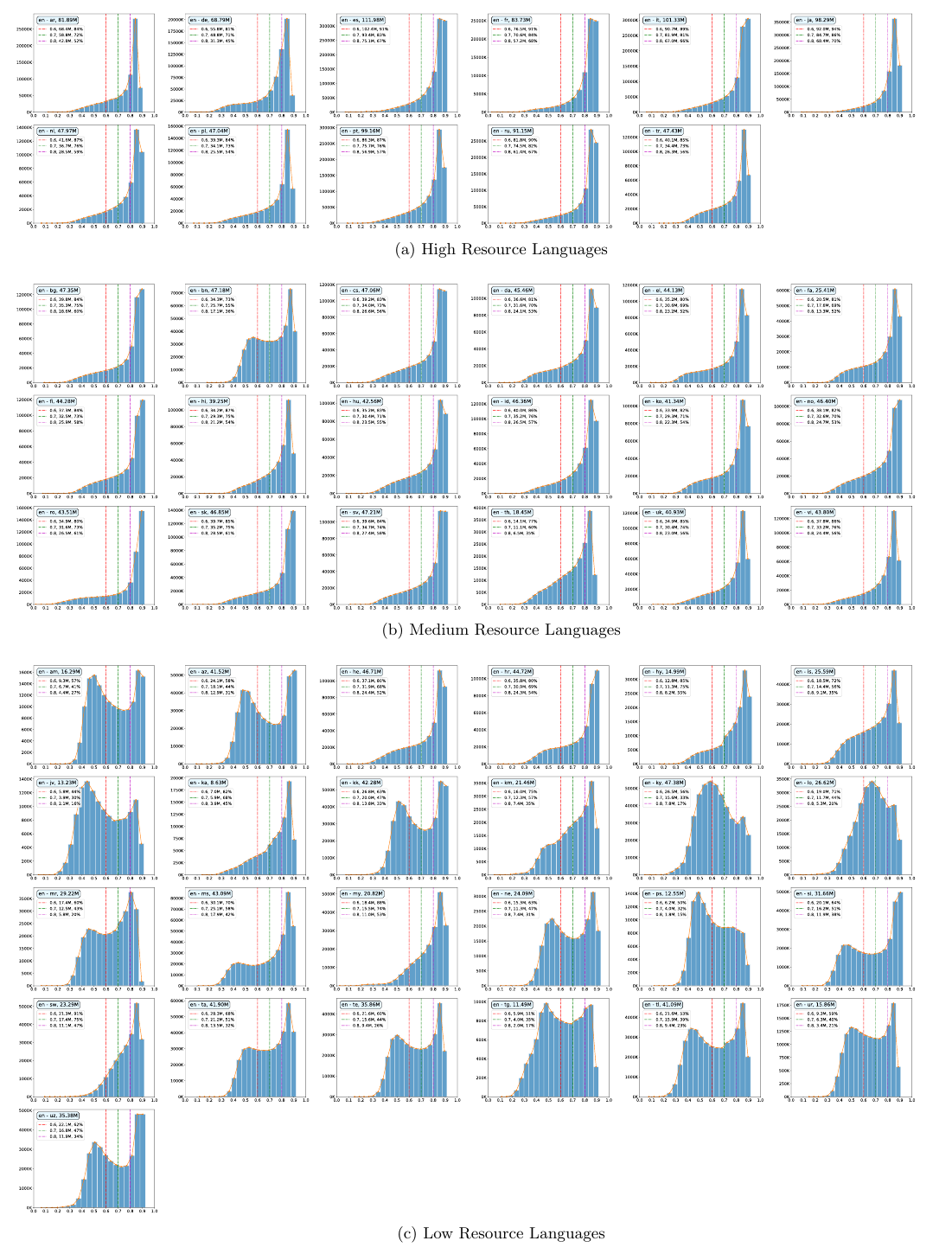}
    \end{adjustbox}
    \caption{COMETKiwi score distributions for bilingual sentence pairs (En-X) are shown as histograms. Vertical lines indicate quality thresholds at 0.6 (red), 0.7 (green), and 0.8 (magenta), with the legend specifying the number and proportion of sentence pairs exceeding each threshold. Some language pairs are excluded due to COMETKiwi's limited language support. }
    \label{fig:en_data}
\end{figure*}
\end{center}

\begin{figure*}[htbp]
    \centering
    \begin{adjustbox}{width=1.05\textwidth, center}
    \includegraphics[width=\linewidth]{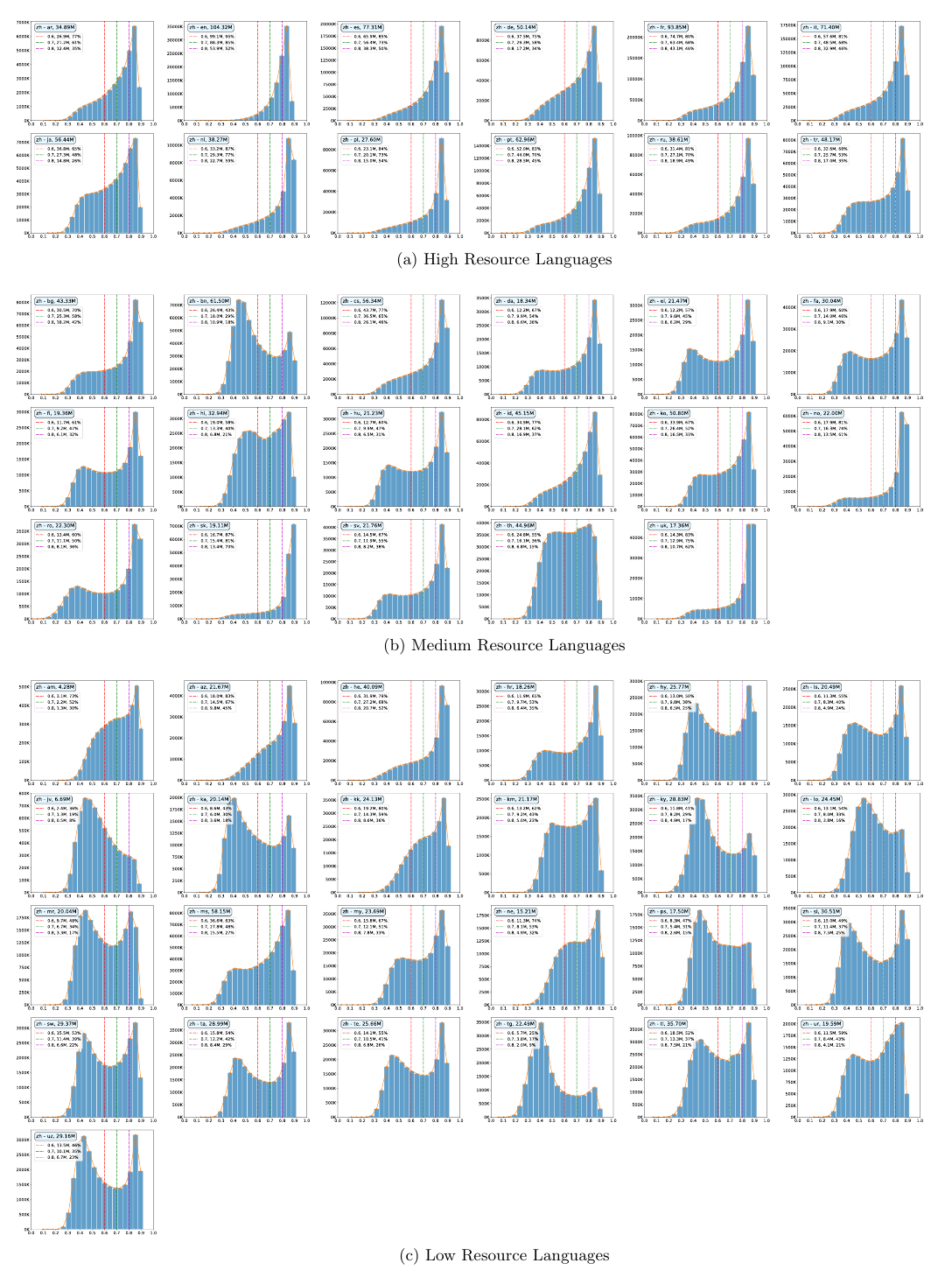}
    \end{adjustbox}
    \caption{COMETKiwi score distributions for bilingual sentence pairs (Zh-X) are shown as histograms. Vertical lines indicate quality thresholds at 0.6 (red), 0.7 (green), and 0.8 (magenta), with the legend specifying the number and proportion of sentence pairs exceeding each threshold. Some language pairs are excluded due to COMETKiwi's limited language support. }
    \label{fig:zh_data}
\end{figure*}

Figures \ref{fig:en_data} and \ref{fig:zh_data} present the COMETKiwi score  \citep{DBLP:conf/wmt/ReiTGZFMSGACLM22} distributions for the English-centric (En $\leftrightarrow$ X) and Chinese-centric (Zh $\leftrightarrow$ X) portions of our curated CPT bilingual corpus, respectively.
A primary observation is that the score distributions for low-resource languages are noticeably skewed to the left compared to the high-resource counterparts. 
This skew is particularly pronounced for the Chinese-centric low-resource pairs (Figure \ref{fig:zh_data}), highlighting the challenge of sourcing non-English-centric data.
We attribute this phenomenon to two factors.
First, it likely reflects the real scarcity of clean, high-quality bilingual data for many low-resource languages, especially those paired with non-English.
Second, it may reveal a bias in the quality estimation (QE) model itself—models like COMETKiwi may yield less reliable scores for underrepresented or non-English language pairs due to limited exposure during training.
This could lead to systematically lower scores for some language pairs, irrespective of their true quality.

Overall, these findings point to a dual challenge in multilingual MT: the limited availability of clean bilingual data for many non-English-centric pairs, and the potential calibration limitations of current QE models in such settings. 
They also motivate future work on QE methods that are better calibrated and more robust across diverse, non-English-centric language pairs.

\section{CPT Gains Across Languages}
\label{sec:appendix_cpt_analysis}

This section details the language-specific impact of Continued Pre-training (CPT), extending the aggregated ablation results in \cref{sec:main_results}.
Figure \ref{fig:compare_cpt} presents COMET score comparisons between Base+SFT+SD (without CPT) and Base+CPT+SFT+SD (with CPT) across four translation directions per language.

Across nearly all languages and directions, incorporating CPT yields consistent performance improvements, underscoring its role as a fundamental adaptation step. 
However, the magnitude of this gain is not uniform, and is more pronounced for medium- and low-resource languages. 
This trend aligns with the expectation that CPT is particularly beneficial for strengthening foundational linguistic knowledge where the base model lacks sufficient prior exposure.
Additionally, for many high- and medium-resource languages, the baseline model exhibits lower performance in the En/Zh~$\to$~X directions than in X~$\to$~En/Zh, consistent with prior observations \citep{DBLP:journals/corr/abs-1907-05019} that generating into diverse target languages is inherently more challenging.
CPT substantially improves performance in these directions, highlighting its effectiveness in strengthening multilingual generation capabilities.

\section{Additional Evaluation}
\label{sec:tier_wmt24}

This appendix complements \cref{sec:comparison} by (i) reporting FLORES results stratified by resource tier, and (ii) evaluating on WMT24++ \citep{deutsch2025wmt24expandinglanguagecoverage} to test robustness beyond FLORES.

\subsection{FLORES Breakdown by Resource Tier}
\cref{tab:compare_mmt_comet} provides a resource-tier breakdown that complements the direction-level averages reported in the main text, and helps assess whether improvements are concentrated in a particular subset of languages. 
In the high-resource tier, where most baselines already perform strongly, LMT-60-4B/8B remains consistently competitive across all four directions and is frequently among the top systems, suggesting that the overall gains are not achieved by sacrificing performance on well-resourced pairs. 
In the medium- and low-resource tiers, the relative advantage of LMT becomes more pronounced and more stable, particularly on Chinese-centric directions where several baselines either lag behind or do not support the full direction set. 
For example, on the 54-language overlap with Aya-101-13B, LMT-60-4B improves low-resource En$\to$X from 81.68 to 86.92 and low-resource X$\to$Zh from 81.58 to 85.98. 
On the 59-language overlap with NLLB-54B, LMT-60-4B yields a clear margin on low-resource X$\to$Zh (80.56$\to$85.88) and improves low-resource Zh$\to$X (82.33$\to$84.52), while remaining strong on En-centric directions. 

Overall, the tiered results indicate that LMT’s performance is broadly competitive on high-resource pairs and delivers larger, more consistent gains on the long tail, rather than being driven by a narrow subset of directions.

\subsection{Results on WMT24++}

\cref{tab:wmt24pp} reports results on WMT24++, a document-level benchmark that emphasizes cross-sentence coherence and discourse-level adequacy.
In this setting, LMT remains competitive on many directions but trails Hunyuan-MT on several subsets.
More broadly, these results are consistent with a general limitation of current LLM-based MT systems: document-level translation is still challenging, likely because most models are post-trained primarily with sentence-level supervision and receive limited exposure to discourse-level signals.
For LMT, our SFT focuses on sentence pairs and includes little document-context training, which can weaken performance on benchmarks that require cross-sentence consistency.
In future work, we will incorporate targeted document-level parallel data and context-aware SFT to strengthen discourse-level behaviors.

\input{appendices/Translation_Error}

\begin{table*}[htbp]
\centering
\begin{tabular}{p{4cm}p{2cm}p{2cm}p{4cm}}
    \hline
    \textbf{Hyperparameter} & \textbf{CPT Stage} & \textbf{SFT Stage} & \textbf{GRPO Stage} \\
    \hline
    Learning Rate & 2e-5 & 2e-5 & 5e-7 \\
    Adam $\beta$ & (0.9, 0.999) & (0.9, 0.999) & (0.9, 0.999) \\
    LR Scheduler & cosine & cosine & cosine \\
    Number of Epochs & 1  & 1  & 1 \\
    Global Batch Size & 1536 & 1024 & 1024 \\
    Max Length & 2048 & 1024 & 1024 \\ 
    Train Steps & 40,000 & 500 & 500 \\
    Warmup Ratio & 0.05 & 0.01 & 0.01 \\
    Weight Decay & 0.01  & 0.01  & 0.01 \\
    Rollout & - & - &  8 \\
    Temperature & - & - & 1.0 \\
    KL $\beta$ & - & - &  0.001 \\
\hline
\end{tabular}
\caption{Hyperparameter configuration during training.}
\label{tab:hyperparameters}
\end{table*}

\definecolor{background_trans}{RGB}{227,229,252}  
\definecolor{background_general}{RGB}{228,242,227}  

\setlength{\fboxsep}{1pt} 
\renewcommand{\arraystretch}{1.2}
\begin{table*}[t!]
  \centering
  \begin{adjustbox}{max width=\textwidth}
  \setlength{\tabcolsep}{3pt}
  \begin{tabular}{clccccccccccccccc}
    \toprule
     \multirow{2.6}{*}{\makecell{\textbf{\# Langs} \\ \textbf{(Hig./Med./Low)}}} & \multirow{2.6}{*}{\textbf{Model}} &\multicolumn{4}{c}{\textbf{High Resource}} &\multicolumn{4}{c}{\textbf{Medium Resource}} &\multicolumn{4}{c}{\textbf{Low Resource}} \\ 
     \cmidrule(lr){3-6} \cmidrule(lr){7-10} \cmidrule(lr){11-14} %
    & &\textbf{{En$\to$X}} &\textbf{{X$\to$En}}
    &\textbf{{Zh$\to$X}} &\textbf{{X$\to$Zh}}
    &\textbf{{En$\to$X}} &\textbf{{X$\to$En}}
    &\textbf{{Zh$\to$X}} &\textbf{{X$\to$Zh}}
    &\textbf{{En$\to$X}} &\textbf{{X$\to$En}}
    &\textbf{{Zh$\to$X}} &\textbf{{X$\to$Zh}} \\
    % \midrule
    \specialrule{0.05em}{0pt}{0pt}
    %%%%%%%%%%%%%%%%%%%%%%%%%%%%%%%%%%%%%%%%%%%
    % \cellcolor{background_trans}
    % &TowerInstruct-7B & 88.35 & 88.31 & 85.28 & 86.04 & 89.42 & 88.21 & 86.51 & 85.11 & - & - & - & - \\
    \cellcolor{background_trans}
    &TowerInstruct-13B & 88.78 & 88.51 & 86.15 & 86.83 & 89.92 & 88.53 & 87.44 & 86.66 & - & - & - & - \\
    \cellcolor{background_trans}
    &LMT-60-4B & \underline{89.14} & \underline{88.56} & \underline{86.91} & \textbf{88.42} & \textbf{90.47} & \textbf{88.68} & \underline{88.49} & \underline{88.28} & - & - & - & - \\
    \cellcolor{background_trans} 
    \multirow{-3}{*}{\makecell{\textbf{10} \\ \textbf{(9/1/0)}}} 
    &LMT-60-8B & \textbf{89.28} & \textbf{88.58} & \textbf{87.14} & \underline{88.39} & \textbf{90.47} & \underline{88.61} & \textbf{88.56} & \textbf{88.43} & - & - & - & -  \\
    % \midrule
    \specialrule{0.05em}{0pt}{0pt}
    %%%%%%%%%%%%%%%%%%%%%%%%%%%%%%%%%%%
    \cellcolor{background_general} 
    &Aya-expanse-8B & 88.60 & 88.16 & 86.44 & 86.32 & 88.82 & 88.50 & 86.19 & 86.13 & \textbf{87.82} & 88.41 & \textbf{84.76} & 85.28 \\
    \cellcolor{background_general} 
    &LMT-60-4B & \underline{89.43} & \underline{88.46} & \underline{87.20} & \textbf{88.19} & \underline{89.67} & \textbf{88.85} & \underline{86.98} & \textbf{87.81} & 87.55 & \underline{88.58} & \underline{84.69} & \underline{87.27} \\
    \cellcolor{background_general} 
    \multirow{-3}{*}{\makecell{\textbf{23} \\ \textbf{(13/9/1)}}} 
    &LMT-60-8B & \textbf{89.60} & \textbf{88.50} & \textbf{87.41} & \underline{88.17} & \textbf{89.87} & \textbf{88.85} & \textbf{87.18} & \underline{87.80} & \underline{87.80} & \textbf{88.65} & 84.41 & \textbf{87.39} \\ 
    % \midrule
    \specialrule{0.05em}{0pt}{0pt}
    %%%%%%%%%%%%%%%%%%%%%%%%%%%%%%%%%%
    \cellcolor{background_trans} 
    &Seed-X-PPO-7B & \textbf{89.91} & \textbf{88.59} & \textbf{87.73} & 87.94 & \textbf{91.58} & \textbf{89.48} & \textbf{89.25} & 88.07 & \textbf{91.27} & \textbf{89.22} & \textbf{88.78} & 87.39 \\
    \cellcolor{background_trans} 
    &LMT-60-4B & 89.43 & 88.46 & 87.20 & \textbf{88.19} & 91.14 & \underline{89.26} & 88.84 & \textbf{88.26} & 91.09 & 89.00 & 88.46 & \underline{87.84} \\
    \cellcolor{background_trans} 
    \multirow{-3}{*}{\makecell{\textbf{27} \\ \textbf{(13/12/2)}}} 
    &LMT-60-8B & \underline{89.60} & \underline{88.50} & \underline{87.41} & \underline{88.17} & \underline{91.26} & 89.24 & \underline{89.04} & \underline{88.20} & 91.14 & \underline{89.01} & \underline{88.67} & \textbf{87.88} \\ 
    % \midrule
    \specialrule{0.05em}{0pt}{0pt}
    %%%%%%%%%%%%%%%%%%%%%%%%%%%%%%%%%%
    %%%%%%%%%%%%%%%%%%%%%%%%%%%%%%%%%%%%%%%%%%%
    % \cellcolor{background_trans} 
    % &GemmaX2-28-2B & 88.64 & 88.18 & 85.92 & 86.89 & 87.97 & 88.76 & 84.60 & 86.52 & 85.23 & 87.85 & 81.45 & 85.06 \\
    \cellcolor{background_trans} 
    &GemmaX2-28-9B & 89.34 & \textbf{88.62} & 86.99 & 87.66 & 88.71 & \textbf{89.46} & 85.82 & 87.59 & 86.42 & \textbf{88.93} & 82.81 & \underline{86.63} \\
    \cellcolor{background_trans} 
    &LMT-60-4B & \underline{89.43} & 88.46 & \underline{87.20} & \textbf{88.19} & \underline{89.11} & 89.04 & \underline{86.16} & \underline{87.99} & \underline{87.06} & 88.04 & \underline{83.63} & 86.56 \\
    \cellcolor{background_trans} 
    \multirow{-3}{*}{\makecell{\textbf{28} \\ \textbf{(13/8/7)}}} 
    &LMT-60-8B & \textbf{89.60} & \underline{88.50} & \textbf{87.41} & \underline{88.17} & \textbf{89.21} & \underline{89.06} & \textbf{86.29} & \textbf{88.03} & \textbf{87.09} & \textbf{88.30} & \textbf{83.73} & \textbf{86.77} \\
    % \midrule
    \specialrule{0.05em}{0pt}{0pt}
    %%%%%%%%%%%%%%%%%%%%%%%%%%%%%%%%%%%%%%
    \cellcolor{background_trans} 
    &Hunyuan-MT-7B & \underline{89.43} & 87.56 & 87.08 & 87.38 & 89.10 & 87.76 & 86.29 & 86.77 & 82.74 & 84.43 & 79.27 & 83.50 \\
    \cellcolor{background_trans} 
    &LMT-60-4B & \underline{89.43} & \underline{88.46} & \underline{87.20} & \textbf{88.19} & \underline{89.28} & \underline{88.89} & \underline{86.49} & \underline{87.91} & \underline{87.67} & \underline{86.96} & \underline{84.63} & \underline{85.96} \\
    \cellcolor{background_trans} 
    \multirow{-3}{*}{\makecell{\textbf{35} \\ \textbf{(13/9/13)}}} 
    &LMT-60-8B & \textbf{89.60} & \textbf{88.50} & \textbf{87.41} & \underline{88.17} & \textbf{89.41} & \textbf{88.90} & \textbf{86.64} & \textbf{87.93} & \textbf{87.75} & \textbf{87.22} & \textbf{84.74} & \textbf{86.22} \\
    % \midrule
    \specialrule{0.05em}{0pt}{0pt}
    %%%%%%%%%%%%%%%%%%%%%%%%%%%%%%%%%%%%%%%%%%% 
    \cellcolor{background_trans} 
    &X-ALMA-13B & 89.41 & \textbf{88.51} & - & - & \underline{90.37} & \textbf{89.29} & - & - & 87.15 & 87.98 & - & - \\
    \cellcolor{background_trans} 
    &LMT-60-4B & \underline{89.43} & 88.46 & - & - & 90.31 & \underline{89.13} & - & - & \underline{87.47} & \underline{88.24} & - & - \\
    \cellcolor{background_trans} 
    \multirow{-3}{*}{\makecell{\textbf{40} \\ \textbf{(13/16/11)}}} 
    &LMT-60-8B & \textbf{89.60} & \underline{88.50} & - & - & \textbf{90.47} & \underline{89.13} & - & - & \textbf{87.55} & \textbf{88.40} & - & - \\ 
    % \midrule
    \specialrule{0.05em}{0pt}{0pt}
    %%%%%%%%%%%%%%%%%%%%%%%%%%%%%%%%%%%%%%%%%%% 
    \cellcolor{background_general} 
    &Aya-101-13B & 87.00 & 86.55 & 84.34 & 83.29 & 87.54 & 87.32 & 84.26 & 83.26 & 81.68 & 85.71 & 77.92 & 81.58 \\
    \cellcolor{background_general} 
    &LMT-60-4B & \underline{89.43} & \underline{88.46} & \underline{87.20} & \textbf{88.19} & \underline{90.23} & \textbf{89.10} & \underline{87.52} & \textbf{87.97} & \underline{86.92} & \underline{87.43} & \underline{83.42} & \underline{85.98} \\
    \cellcolor{background_general} 
    \multirow{-3}{*}{\makecell{\textbf{54} \\ \textbf{(13/18/23)}}} 
    &LMT-60-8B & \textbf{89.60} & \textbf{88.50} & \textbf{87.41} & \underline{88.17} & \textbf{90.39} & \textbf{89.10} & \textbf{87.70} & \underline{87.95} & \textbf{87.01} & \textbf{87.72} & \textbf{83.56} & \textbf{86.20} \\
    % \midrule
    \specialrule{0.05em}{0pt}{0pt}
    %%%%%%%%%%%%%%%%%%%%%%%%%%%%%%%%%%%%%%%%%%%
    \cellcolor{background_general}
    &LLaMAX3-Alpaca & 85.22 & 87.19 & 82.28 & 82.49 & 84.80 & 87.87 & 80.90 & 82.23 & 76.69 & 84.60 & 71.97 & 79.38 \\
    \cellcolor{background_general}
    &LMT-60-4B & \underline{89.43} & \underline{88.46} & \underline{87.20} & \textbf{88.19} & \underline{90.23} & \textbf{89.10} & \underline{87.52} & \textbf{87.97} & \underline{87.10} & \underline{87.40} & \underline{83.69} & \underline{86.03} \\
    \cellcolor{background_general} 
    \multirow{-3}{*}{\makecell{\textbf{55} \\ \textbf{(13/18/24)}}} 
    &LMT-60-8B & \textbf{89.60} & \textbf{88.50} & \textbf{87.41} & \underline{88.17} & \textbf{90.39} & \textbf{89.10} & \textbf{87.70} & \underline{87.95} & \textbf{87.17} & \textbf{87.65} & \textbf{83.82} & \textbf{86.22} \\ 
    % \midrule
    \specialrule{0.05em}{0pt}{0pt}
    %%%%%%%%%%%%%%%%%%%%%%%%%%%%%%%%%%%%%%%%%%%
    % \cellcolor{background_trans} 
    % &NLLB-3.3B & 87.63 & 87.36 & 84.65 & 76.48 & 88.32 & 88.02 & 84.09 & 77.78 & 84.23 & 85.45 & 80.60 & 77.36 \\
    \cellcolor{background_trans} 
    &NLLB-54B & 87.95 & 88.17 & 85.82 & 80.06 & 88.95 & 88.85 & 85.58 & 80.69 & 85.12 & \underline{86.81} & 82.33 & 80.56 \\
    \cellcolor{background_trans} 
    &LMT-60-4B & \underline{89.43} & \underline{88.46} & \underline{87.20} & \textbf{88.19} & \underline{90.23} & \textbf{89.10} & \underline{87.52} & \textbf{87.97} & \underline{87.57} & \underline{86.96} & \underline{84.52} & \underline{85.88} \\
    \cellcolor{background_trans} 
    \multirow{-3}{*}{\makecell{\textbf{59} \\ \textbf{(13/18/28)}}} 
    &LMT-60-8B & \textbf{89.60} & \textbf{88.50} & \textbf{87.41} & \underline{88.17} & \textbf{90.39} & \textbf{89.10} & \textbf{87.70} & \underline{87.95} & \textbf{87.65} & \textbf{87.23} & \textbf{84.64} & \textbf{86.12} \\
    % \bottomrule
    \specialrule{0.08em}{0pt}{0pt}
  \end{tabular}
  \end{adjustbox}
  \caption{
COMET-22 scores of our LMT models compared with a range of \colorbox{background_general}{general-purpose multilingual LLMs} and \colorbox{background_trans}{dedicated MMT models}. Evaluation is conducted only on the intersection of language pairs supported by each baseline and our models. The first column (\# Langs) denotes the number of overlapping languages, followed by their distribution across resource tier (high/medium/low).
\textbf{Bold} numbers indicate the best in each group, and the \underline{underlined} numbers the second best.
The symbol '-' indicates directions not supported by the baseline model.
}
  \label{tab:compare_mmt_comet}
\end{table*}
\renewcommand{\arraystretch}{1}

\definecolor{background_trans}{RGB}{227,229,252}  
\definecolor{background_general}{RGB}{228,242,227}  

\setlength{\fboxsep}{1pt} 
\renewcommand{\arraystretch}{1.2}
\begin{table*}[htbp]
  \centering
  \begin{adjustbox}{max width=\textwidth}
  \setlength{\tabcolsep}{3pt}
  \begin{tabular}{clccccccccccccccc}
    \toprule
     \multirow{2.6}{*}{\makecell{\textbf{\# Langs} \\ \textbf{(Hig./Med./Low)}}} & \multirow{2.6}{*}{\textbf{Model}} &\multicolumn{4}{c}{\textbf{High Resource}} &\multicolumn{4}{c}{\textbf{Medium Resource}} &\multicolumn{4}{c}{\textbf{Low Resource}} \\ 
     \cmidrule(lr){3-6} \cmidrule(lr){7-10} \cmidrule(lr){11-14} %
    & &\textbf{{En$\to$X}} &\textbf{{X$\to$En}}
    &\textbf{{Zh$\to$X}} &\textbf{{X$\to$Zh}}
    &\textbf{{En$\to$X}} &\textbf{{X$\to$En}}
    &\textbf{{Zh$\to$X}} &\textbf{{X$\to$Zh}}
    &\textbf{{En$\to$X}} &\textbf{{X$\to$En}}
    &\textbf{{Zh$\to$X}} &\textbf{{X$\to$Zh}} \\
    % \midrule
    \specialrule{0.05em}{0pt}{0pt}
    %%%%%%%%%%%%%%%%%%%%%%%%%%%%%%%%%%%%%%%%%%%
    \cellcolor{background_trans}
    &TowerInstruct-13B & \underline{37.46} & \textbf{39.55} & 21.56 & 36.72 & \underline{30.54} & \textbf{31.99} & 21.34 & 33.86 & - & - & - & - \\
    \cellcolor{background_trans}
    &LMT-60-4B & 37.35 & 39.15 & \underline{23.03} & \underline{41.93} & 29.38 & 30.95 & \underline{23.20} & \underline{38.59} & - & - & - & -  \\
    \cellcolor{background_trans} 
    \multirow{-3}{*}{\makecell{\textbf{10} \\ \textbf{(9/1/0)}}} 
    &LMT-60-8B & \textbf{38.34} & \underline{39.28} & \textbf{24.25} & \textbf{42.09} & \textbf{31.36} & \underline{31.66} & \textbf{24.20} & \textbf{39.34} & - & - & - & -  \\
    % \midrule
    \specialrule{0.05em}{0pt}{0pt}
    %%%%%%%%%%%%%%%%%%%%%%%%%%%%%%%%%%%
    \cellcolor{background_general} 
    &Aya-expanse-8B & 31.74 & 36.64 & 19.70 & 34.58 & 31.14 & 37.67 & 19.61 & 34.53 & \textbf{27.45} & 42.71 & 13.53 & 35.12 \\
    \cellcolor{background_general} 
    &LMT-60-4B & \underline{33.79} & \underline{37.97} & \underline{21.42} & \underline{41.10} & \underline{32.61} & \underline{39.69} & \underline{21.10} & \underline{40.48} & 24.68 & \underline{44.13} & \underline{14.13} & \underline{41.32} \\
    \cellcolor{background_general} 
    \multirow{-3}{*}{\makecell{\textbf{23} \\ \textbf{(13/9/1)}}} 
    &LMT-60-8B & \textbf{35.00} & \textbf{38.16} & \textbf{22.48} & \textbf{41.33} & \textbf{34.01} & \textbf{39.89} & \textbf{22.00} & \textbf{40.80} & \underline{27.15} & \textbf{44.58} & \textbf{14.53} & \textbf{41.76} \\ 
    % \midrule
    \specialrule{0.05em}{0pt}{0pt}
    %%%%%%%%%%%%%%%%%%%%%%%%%%%%%%%%%%
    \cellcolor{background_trans} 
    &Seed-X-PPO-7B & \textbf{37.88} & 37.14 & \textbf{23.13} & 34.03 & \textbf{38.33} & 40.63 & \textbf{23.12} & 34.95 & \textbf{39.32} & 42.53 & \textbf{21.84} & 33.86 \\
    \cellcolor{background_trans} 
    &LMT-60-4B & 33.79 & 37.97 & 21.42 & \underline{41.10} & 32.83 & \underline{40.77} & 21.38 & \textbf{41.23} & 35.02 & \underline{42.70} & 20.64 & \underline{41.25} \\
    \cellcolor{background_trans} 
    \multirow{-3}{*}{\makecell{\textbf{27} \\ \textbf{(13/12/2)}}} 
    &LMT-60-8B & \underline{35.00} & \textbf{38.16} & \underline{22.48} & \textbf{41.33} & \underline{33.91} & \textbf{40.79} & \underline{22.32} & \underline{41.21} & \underline{35.79} & \textbf{42.72} & \underline{21.52} & \textbf{41.26} \\ 
    % \midrule
    \specialrule{0.05em}{0pt}{0pt}
    %%%%%%%%%%%%%%%%%%%%%%%%%%%%%%%%%%
    %%%%%%%%%%%%%%%%%%%%%%%%%%%%%%%%%%%%%%%%%%%
    \cellcolor{background_trans} 
    &GemmaX2-28-9B & \textbf{36.23} & \textbf{39.78} & \textbf{22.60} & 40.42 & \textbf{32.21} & \textbf{41.59} & \textbf{20.40} & 39.21 & \underline{22.06} & \textbf{43.09} & \textbf{13.74} & \underline{38.21} \\
    \cellcolor{background_trans} 
    &LMT-60-4B & 33.79 & 37.97 & 21.42 & \underline{41.10} & 28.96 & 37.87 & 19.21 & \underline{39.66} & 21.51 & 38.61 & 13.01 & 37.71 \\
    \cellcolor{background_trans} 
    \multirow{-3}{*}{\makecell{\textbf{28} \\ \textbf{(13/8/7)}}} 
    &LMT-60-8B & \underline{35.00} & \underline{38.16} & \underline{22.48} & \textbf{41.33} & \underline{30.22} & \underline{38.22} & \underline{20.12} & \textbf{40.17} & \textbf{22.28} & \underline{39.51} & \underline{13.60} & \textbf{38.42} \\
    % \midrule
    \specialrule{0.05em}{0pt}{0pt}
    %%%%%%%%%%%%%%%%%%%%%%%%%%%%%%%%%%%%%%
    \cellcolor{background_trans} 
    &Hunyuan-MT-7B & 27.72 & 28.65 & 17.55 & 28.49 & 21.24 & 27.38 & 14.27 & 26.48 & 9.99 & 21.81 & 6.59 & 22.04 \\
    \cellcolor{background_trans} 
    &LMT-60-4B & \underline{33.79} & \underline{37.97} & \underline{21.42} & \underline{41.10} & \underline{28.92} & \underline{38.12} & \underline{18.95} & \underline{39.84} & \underline{18.00} & \underline{34.70} & \underline{10.97} & \underline{36.93} \\
    \cellcolor{background_trans} 
    \multirow{-3}{*}{\makecell{\textbf{35} \\ \textbf{(13/9/13)}}} 
    &LMT-60-8B & \textbf{35.00} & \textbf{38.16} & \textbf{22.48} & \textbf{41.33} & \textbf{30.22} & \textbf{38.47} & \textbf{19.88} & \textbf{40.27} & \textbf{19.01} & \textbf{35.74} & \textbf{11.80} & \textbf{37.45} \\
    % \midrule
    \specialrule{0.05em}{0pt}{0pt}
    %%%%%%%%%%%%%%%%%%%%%%%%%%%%%%%%%%%%%%%%%%% 
    \cellcolor{background_trans} 
    &X-ALMA-13B & \textbf{35.76} & \textbf{38.22} & - & - & \textbf{35.12} & \textbf{41.53} & - & - & \textbf{22.16} & \underline{35.74} & - & - \\
    \cellcolor{background_trans} 
    &LMT-60-4B & 33.79 & 37.97 & - & - & 31.87 & 40.42 & - & - & 21.17 & 35.38 & - & -  \\
    \cellcolor{background_trans} 
    \multirow{-3}{*}{\makecell{\textbf{40} \\ \textbf{(13/16/11)}}} 
    &LMT-60-8B & \underline{35.00} & \underline{38.16} & - & - & \underline{33.01} & \underline{40.49} & - & - & \underline{21.96} & \textbf{36.23} & - & - \\ 
    % \midrule
    \specialrule{0.05em}{0pt}{0pt}
    %%%%%%%%%%%%%%%%%%%%%%%%%%%%%%%%%%%%%%%%%%% 
    \cellcolor{background_general} 
    &Aya-101-13B & 22.75 & 30.67 & 12.86 & 24.10 & 20.02 & 32.61 & 10.33 & 24.01 & 7.70 & 29.01 & 3.71 & 21.98 \\
    \cellcolor{background_general} 
    &LMT-60-4B & \underline{33.79} & \underline{37.97} & \underline{21.42} & \underline{41.10} & \underline{30.96} & \underline{40.11} & \underline{19.13} & \underline{40.61} & \underline{20.05} & \underline{36.17} & \underline{11.98} & \underline{37.02} \\
    \cellcolor{background_general} 
    \multirow{-3}{*}{\makecell{\textbf{54} \\ \textbf{(13/18/23)}}} 
    &LMT-60-8B & \textbf{35.00} & \textbf{38.16} & \textbf{22.48} & \textbf{41.33} & \textbf{32.11} & \textbf{40.22} & \textbf{20.08} & \textbf{40.77} & \textbf{20.85} & \textbf{37.26} & \textbf{12.53} & \textbf{37.77}  \\
    % \midrule
    \specialrule{0.05em}{0pt}{0pt}
    %%%%%%%%%%%%%%%%%%%%%%%%%%%%%%%%%%%%%%%%%%%
    \cellcolor{background_general}
    &LLaMAX3-Alpaca & 24.51 & 32.28 & 12.94 & 26.18 & 22.53 & 34.46 & 9.41 & 25.61 & 10.96 & 27.51 & 4.57 & 20.67 \\
    \cellcolor{background_general}
    &LMT-60-4B & \underline{33.79} & \underline{37.97} & \underline{21.42} & \underline{41.10} & \underline{30.96} & \underline{40.11} & \underline{19.13} & \underline{40.61} & \underline{20.38} & \underline{36.01} & \underline{12.27} & \underline{37.10} \\
    \cellcolor{background_general} 
    \multirow{-3}{*}{\makecell{\textbf{55} \\ \textbf{(13/18/24)}}} 
    &LMT-60-8B & \textbf{35.00} & \textbf{38.16} & \textbf{22.48} & \textbf{41.33} & \textbf{32.11} & \textbf{40.22} & \textbf{20.08} & \textbf{40.77} & \textbf{21.11} & \textbf{37.01} & \textbf{12.82} & \textbf{37.80} \\ 
    % \midrule
    \specialrule{0.05em}{0pt}{0pt}
    %%%%%%%%%%%%%%%%%%%%%%%%%%%%%%%%%%%%%%%%%%%
    \cellcolor{background_trans} 
    &NLLB-54B & 33.02 & \textbf{39.15} & 20.81 & 24.51 & \underline{31.67} & \textbf{41.60} & 18.40 & 24.96 & \underline{19.20} & \textbf{37.06} & 10.79 & 25.04 \\
    \cellcolor{background_trans} 
    &LMT-60-4B & \underline{33.79} & 37.97 & \underline{21.42} & \underline{41.10} & 30.96 & 40.11 & \underline{19.13} & \underline{40.61} & 18.93 & 34.77 & \underline{11.56} & \underline{36.79} \\
    \cellcolor{background_trans} 
    \multirow{-3}{*}{\makecell{\textbf{59} \\ \textbf{(13/18/28)}}} 
    &LMT-60-8B & \textbf{35.00} & \underline{38.16} & \textbf{22.48} & \textbf{41.33} & \textbf{32.11} & \underline{40.22} & \textbf{20.08} & \textbf{40.77} & \textbf{19.75} & \underline{35.81} & \textbf{12.21} & \textbf{37.47} \\
    % \bottomrule
    \specialrule{0.08em}{0pt}{0pt}
  \end{tabular}
  \end{adjustbox}
  \caption{
BLEU scores of our LMT models compared with a range of \colorbox{background_general}{general-purpose multilingual LLMs} and \colorbox{background_trans}{dedicated MMT models}. Evaluation is conducted only on the intersection of language pairs supported by each baseline and our models. The first column (\# Langs) denotes the number of overlapping languages, followed by their distribution across resource tier (high/medium/low).
\textbf{Bold} numbers indicate the best in each group, and the \underline{underlined} numbers the second best.
The symbol '-' indicates directions not supported by the baseline model.
}
  \label{tab:main_result2}
\end{table*}
\renewcommand{\arraystretch}{1}

% \definecolor{background_color1}{RGB}{255,229,228}
% \definecolor{background_color2}{RGB}{227,229,252}
% \definecolor{background_color3}{RGB}{228,242,227}
\definecolor{myorange}{RGB}{242,181,111} % 橙
\definecolor{myblue}{RGB}{113,183,237} % 蓝
\setlength{\fboxsep}{1pt} % 设置colorbox文字与框的间距
\begin{figure*}[h]
    \centering
    \begin{adjustbox}{width=1.1\linewidth, center}
    \includegraphics[width=1.1\linewidth]{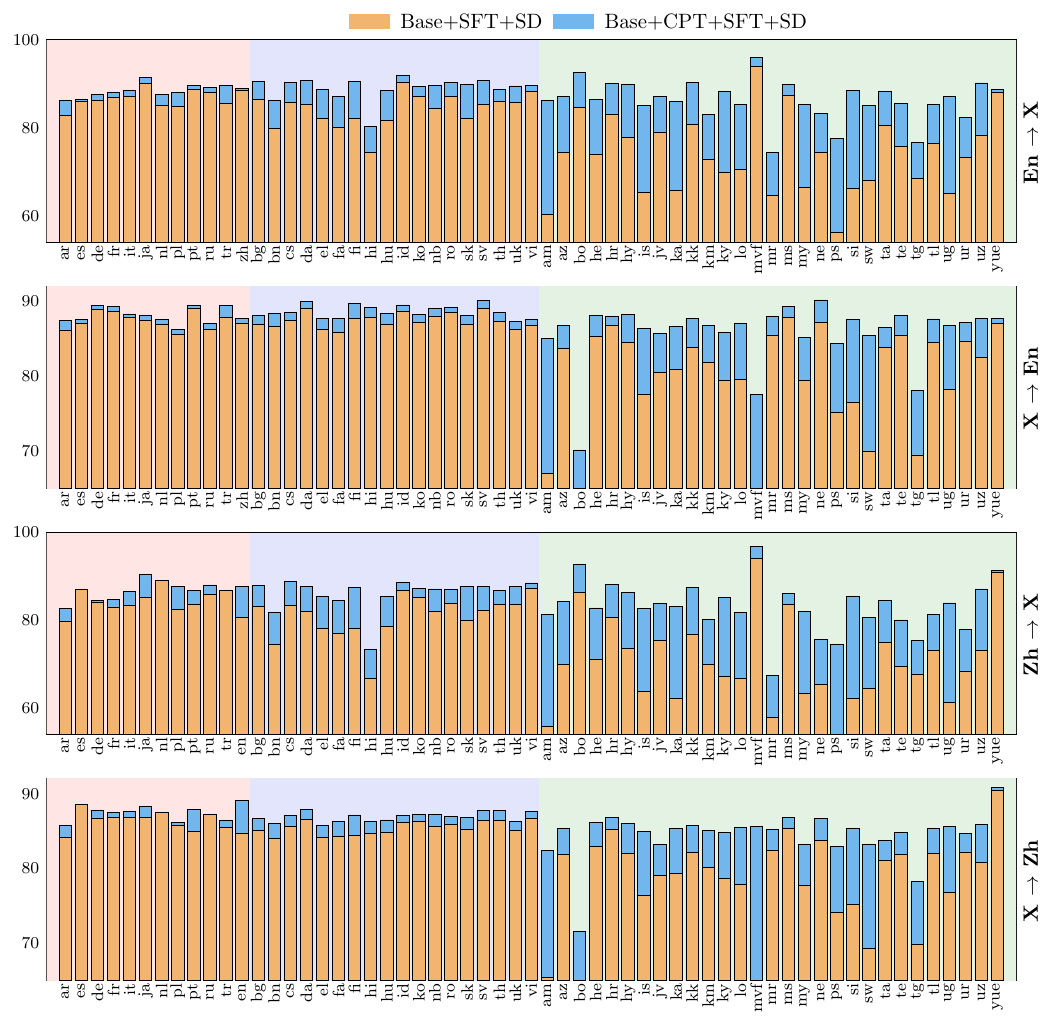}
    \end{adjustbox}
    \caption{
    Performance improvements brought by Continued Pre-training (CPT).
    Languages are grouped by resource level: \colorbox{background_color1}{high}, \colorbox{background_color2}{medium }, and \colorbox{background_color3}{low}.
    The \colorbox{myorange}{orange portion} of the bar shows the performance of the model without CPT (Base+SFT+SD), while the total height of the bar represents the performance after including the CPT stage (Base+CPT+SFT+SD). The \colorbox{myblue}{blue portion} visually represents the performance gain ($\Delta$COMET) contributed by CPT.}
    \label{fig:compare_cpt}
\end{figure*}

\newpage

\renewcommand{\arraystretch}{1.2}
\begin{table*}[h]
  \centering
  \begin{adjustbox}{max width=\textwidth}
  \setlength{\tabcolsep}{4pt}
  \begin{tabular}{clccccccccccccccc}
    \toprule
     \multirow{2.5}{*}{\makecell{\textbf{\# Langs} \\ \textbf{(Hig./Med./Low)}}} & \multirow{2.5}{*}{\textbf{Model}} &\multicolumn{4}{c}{\textbf{High Resource}} &\multicolumn{4}{c}{\textbf{Medium Resource}} &\multicolumn{4}{c}{\textbf{Low Resource}} \\ 
     \cmidrule(lr){3-6} \cmidrule(lr){7-10} \cmidrule(lr){11-14} %
    & &\textbf{{En$\to$X}} &\textbf{{X$\to$En}}
    &\textbf{{Zh$\to$X}} &\textbf{{X$\to$Zh}}
    &\textbf{{En$\to$X}} &\textbf{{X$\to$En}}
    &\textbf{{Zh$\to$X}} &\textbf{{X$\to$Zh}}
    &\textbf{{En$\to$X}} &\textbf{{X$\to$En}}
    &\textbf{{Zh$\to$X}} &\textbf{{X$\to$Zh}} \\
    % \midrule
    \specialrule{0.05em}{0pt}{0pt}
    %%%%%%%%%%%%%%%%%%%%%%%%%%%%%%%%%%%%%%%%%%%
    \cellcolor{background_trans} 
    &TowerInstruct-13B & 82.27 & 83.33 & 81.78 & 81.06 & 84.28 & 81.29 & 82.14 & 78.12 & - & - & - & - \\
    \cellcolor{background_trans} 
    &LMT-60-4B & 82.64 & 83.37 & 82.66 & 85.09 & 86.92 & 84.35 & 85.82 & 84.96 & - & - & - & - \\
    \cellcolor{background_trans} 
    \multirow{-3}{*}{\makecell{\textbf{10} \\ \textbf{(9/1/0)}}} 
    &LMT-60-8B & 83.04 & 83.54 & 83.22 & 85.01 & 87.39 & 84.97 & 86.73 & 85.29 & - & - & - & - \\
    % \midrule
    \specialrule{0.05em}{0pt}{0pt}
    %%%%%%%%%%%%%%%%%%%%%%%%%%%%%%%%%%%%%%
    \cellcolor{background_general} 
    &Aya-expanse-8B & 82.50 & 83.03 & 82.85 & 82.08 & 83.38 & 83.78 & 83.38 & 82.10 & 82.92 & 83.46 & 81.59 & 80.98 \\
    \cellcolor{background_general} 
    &LMT-60-4B & 83.11 & 82.93 & 82.79 & 84.46 & 83.97 & 82.95 & 83.06 & 83.57 & 80.95 & 83.14 & 77.72 & 83.39 \\
    \cellcolor{background_general} 
    \multirow{-3}{*}{\makecell{\textbf{23} \\ \textbf{(13/9/1)}}} 
    &LMT-60-8B & 83.41 & 83.15 & 83.32 & 84.59 & 84.12 & 83.84 & 83.72 & 84.36 & 81.34 & 83.47 & 79.18 & 83.80 \\
    % \midrule
    \specialrule{0.05em}{0pt}{0pt}
    %%%%%%%%%%%%%%%%%%%%%%%%%%%%%%%%%%%%%%
    \cellcolor{background_trans} 
    &GemmaX2-28-9B & 75.49 & 76.42 & 74.40 & 76.26 & 74.92 & 68.62 & 75.34 & 75.97 & 75.04 & 58.88 & 73.21 & 76.92 \\
    \cellcolor{background_trans} 
    &LMT-60-4B & 83.11 & 82.93 & 82.79 & 84.46 & 83.64 & 83.40 & 82.68 & 83.67 & 81.30 & 82.92 & 78.80 & 82.78 \\
    \cellcolor{background_trans} 
    \multirow{-3}{*}{\makecell{\textbf{24} \\ \textbf{(13/8/3)}}} 
    &LMT-60-8B & 83.41 & 83.15 & 83.32 & 84.59 & 83.77 & 84.18 & 83.02 & 84.76 & 81.27 & 83.60 & 79.49 & 83.65 \\
    % \midrule
    \specialrule{0.05em}{0pt}{0pt}
    %%%%%%%%%%%%%%%%%%%%%%%%%%%%%%%%%%%%%%
    \cellcolor{background_trans} 
    &Hunyuan-MT-7B & 85.35 & 83.32 & 84.21 & 85.30 & 86.59 & 84.40 & 84.55 & 85.11 & 81.29 & 83.58 & 77.83 & 83.23 \\
    \cellcolor{background_trans} 
    &LMT-60-4B & 83.11 & 82.93 & 82.79 & 84.46 & 83.74 & 83.14 & 82.77 & 83.54 & 79.74 & 81.92 & 76.81 & 80.70 \\
    \cellcolor{background_trans} 
    \multirow{-3}{*}{\makecell{\textbf{28} \\ \textbf{(13/9/6)}}} 
    &LMT-60-8B & 83.41 & 83.15 & 83.32 & 84.59 & 83.96 & 83.98 & 83.22 & 84.60 & 80.83 & 83.56 & 78.24 & 82.93 \\
    % \midrule
    \specialrule{0.05em}{0pt}{0pt}
    %%%%%%%%%%%%%%%%%%%%%%%%%%%%%%%%%%%%%%%%%%% 
    \cellcolor{background_trans} 
    &X-ALMA-13B & 80.15 & 79.31 & - & - & 81.60 & 78.87 & - & - & 75.35 & 75.98 & - & - \\
    \cellcolor{background_trans} 
    &LMT-60-4B & 83.11 & 82.93 & - & - & 84.69 & 83.30 & - & - & 78.41 & 82.82 & - & - \\
    \cellcolor{background_trans} 
    \multirow{-3}{*}{\makecell{\textbf{33} \\ \textbf{(13/16/4)}}} 
    &LMT-60-8B & 83.41 & 83.15 & - & - & 84.96 & 84.10 & - & - & 78.78 & 83.72 & - & - \\
    % \bottomrule
    \specialrule{0.05em}{0pt}{0pt}
    %%%%%%%%%%%%%%%%%%%%%%%%%%%%%%%%%%%%%%%%%%% 
    \cellcolor{background_general} 
    &Aya-101-13B & 75.14 & 77.68 & 77.85 & 75.00 & 76.64 & 78.69 & 77.94 & 75.06 & 71.80 & 77.87 & 70.82 & 74.04 \\
    \cellcolor{background_general} 
    &LMT-60-4B & 83.11 & 82.93 & 82.79 & 84.46 & 84.62 & 83.25 & 83.50 & 83.76 & 80.49 & 81.76 & 77.99 & 81.04 \\
    \cellcolor{background_general} 
    \multirow{-3}{*}{\makecell{\textbf{39} \\ \textbf{(13/18/8)}}} 
    &LMT-60-8B & 83.41 & 83.15 & 83.32 & 84.59 & 84.88 & 84.04 & 84.07 & 84.40 & 81.36 & 83.23 & 79.11 & 82.94 \\
    % \midrule
    \specialrule{0.05em}{0pt}{0pt}
    %%%%%%%%%%%%%%%%%%%%%%%%%%%%%%%%%%%%%%
  \end{tabular}
  \end{adjustbox}
  \caption{COMET-22 results on the WMT24++ benchmark (document-level). Only models without WMT24++ training are included. Evaluation is conducted only on the intersection of language pairs supported by each baseline and our models. The first column (\# Langs) denotes the number of overlapping languages, followed by their distribution across resource tier (high/medium/low).
The symbol '-' indicates directions not supported by the baseline model.}
  \label{tab:wmt24pp}
\end{table*}
\renewcommand{\arraystretch}{1} 

\renewcommand{\arraystretch}{1.2}
\begin{table*}[h]
  \centering
  \begin{adjustbox}{max width=\textwidth}
  \setlength{\tabcolsep}{4pt}
  \begin{tabular}{clccccccccccccccc}
    \toprule
     \multirow{2.5}{*}{\makecell{\textbf{\# Langs} \\ \textbf{(Hig./Med./Low)}}} & \multirow{2.5}{*}{\textbf{Model}} &\multicolumn{4}{c}{\textbf{High Resource}} &\multicolumn{4}{c}{\textbf{Medium Resource}} &\multicolumn{4}{c}{\textbf{Low Resource}} \\ 
     \cmidrule(lr){3-6} \cmidrule(lr){7-10} \cmidrule(lr){11-14} %
    & &\textbf{{En$\to$X}} &\textbf{{X$\to$En}}
    &\textbf{{Zh$\to$X}} &\textbf{{X$\to$Zh}}
    &\textbf{{En$\to$X}} &\textbf{{X$\to$En}}
    &\textbf{{Zh$\to$X}} &\textbf{{X$\to$Zh}}
    &\textbf{{En$\to$X}} &\textbf{{X$\to$En}}
    &\textbf{{Zh$\to$X}} &\textbf{{X$\to$Zh}} \\
    % \midrule
    \specialrule{0.05em}{0pt}{0pt}
    %%%%%%%%%%%%%%%%%%%%%%%%%%%%%%%%%%%%%%%%%%%
    \cellcolor{background_trans} 
    &TowerInstruct-13B & 32.33 & 35.15 & 15.54 & 25.74 & 16.07 & 16.92 & 10.81 & 15.26 & - & - & - & - \\
    \cellcolor{background_trans} 
    &LMT-60-4B & 28.89 & 28.59 & 17.41 & 34.83 & 15.39 & 17.10 & 18.43 & 30.19 & - & - & - & - \\
    \cellcolor{background_trans} 
    \multirow{-3}{*}{\makecell{\textbf{10} \\ \textbf{(9/1/0)}}} 
    &LMT-60-8B & 32.45 & 34.24 & 19.88 & 29.31 & 25.28 & 21.57 & 22.41 & 24.05 & - & - & - & - \\
    % \midrule
    \specialrule{0.05em}{0pt}{0pt}
    %%%%%%%%%%%%%%%%%%%%%%%%%%%%%%%%%%%%%%
    \cellcolor{background_general} 
    &Aya-expanse-8B & 28.46 & 32.20 & 17.15 & 31.30 & 27.15 & 31.95 & 16.36 & 30.07 & 24.48 & 34.64 & 13.12 & 29.58 \\
    \cellcolor{background_general} 
    &LMT-60-4B & 25.00 & 26.40 & 15.46 & 33.16 & 19.33 & 24.47 & 14.12 & 27.48 & 15.28 & 30.02 & 8.12 & 30.96 \\
    \cellcolor{background_general} 
    \multirow{-3}{*}{\makecell{\textbf{23} \\ \textbf{(13/9/1)}}} 
    &LMT-60-8B & 28.18 & 29.74 & 17.48 & 28.18 & 26.58 & 21.72 & 17.17 & 26.58 & 15.78 & 35.22 & 8.12 & 29.73 \\
    % \midrule
    \specialrule{0.05em}{0pt}{0pt}
    %%%%%%%%%%%%%%%%%%%%%%%%%%%%%%%%%%%%%%
    \cellcolor{background_trans} 
    &GemmaX2-28-9B & 16.84 & 22.03 & 10.56 & 21.98 & 12.39 & 13.60 & 8.06 & 16.88 & 12.60 & 9.60 & 7.19 & 19.33 \\
    \cellcolor{background_trans} 
    &LMT-60-4B & 25.00 & 26.40 & 15.46 & 33.16 & 15.27 & 21.77 & 12.03 & 25.79 & 20.41 & 25.37 & 11.43 & 27.58 \\
    \cellcolor{background_trans} 
    \multirow{-3}{*}{\makecell{\textbf{24} \\ \textbf{(13/8/3)}}} 
    &LMT-60-8B & 28.18 & 29.74 & 17.48 & 28.18 & 20.92 & 18.76 & 14.53 & 26.57 & 20.81 & 27.36 & 12.06 & 29.42 \\
    % \midrule
    \specialrule{0.05em}{0pt}{0pt}
    %%%%%%%%%%%%%%%%%%%%%%%%%%%%%%%%%%%%%%
    \cellcolor{background_trans} 
    &Hunyuan-MT-7B & 19.28 & 26.60 & 14.40 & 28.44 & 14.66 & 23.38 & 11.12 & 25.21 & 7.47 & 22.09 & 5.51 & 20.84 \\
    \cellcolor{background_trans} 
    &LMT-60-4B & 25.00 & 26.40 & 15.46 & 33.16 & 15.72 & 22.22 & 11.96 & 25.81 & 14.13 & 20.09 & 7.80 & 21.27 \\
    \cellcolor{background_trans} 
    \multirow{-3}{*}{\makecell{\textbf{28} \\ \textbf{(13/9/6)}}} 
    &LMT-60-8B & 28.18 & 29.74 & 17.48 & 28.18 & 21.40 & 18.74 & 14.56 & 26.30 & 16.29 & 18.54 & 9.33 & 23.59 \\
    % \midrule
    \specialrule{0.05em}{0pt}{0pt}
    %%%%%%%%%%%%%%%%%%%%%%%%%%%%%%%%%%%%%%%%%%% 
    \cellcolor{background_trans} 
    &X-ALMA-13B & 11.98 & 12.58 & - & - & 11.99 & 11.55 & - & - & 6.72 & 9.07 & - & - \\
    \cellcolor{background_trans} 
    &LMT-60-4B & 25.00 & 26.40 & - & - & 21.24 & 26.08 & - & - & 14.84 & 23.11 & - & - \\
    \cellcolor{background_trans} 
    \multirow{-3}{*}{\makecell{\textbf{33} \\ \textbf{(13/16/4)}}} 
    &LMT-60-8B & 28.18 & 29.74 & - & - & 28.10 & 26.45 & - & - & 17.00 & 22.03 & - & - \\
    % \bottomrule
    \specialrule{0.05em}{0pt}{0pt}
    %%%%%%%%%%%%%%%%%%%%%%%%%%%%%%%%%%%%%%%%%%% 
    \cellcolor{background_general} 
    &Aya-101-13B & 10.26 & 15.01 & 5.91 & 7.53 & 11.39 & 15.75 & 5.11 & 8.96 & 6.63 & 15.39 & 2.40 & 8.61 \\
    \cellcolor{background_general} 
    &LMT-60-4B & 25.00 & 26.40 & 15.46 & 33.16 & 20.08 & 25.17 & 13.41 & 28.36 & 15.73 & 19.76 & 8.90 & 22.39 \\
    \cellcolor{background_general} 
    \multirow{-3}{*}{\makecell{\textbf{39} \\ \textbf{(13/18/8)}}} 
    &LMT-60-8B & 28.18 & 29.74 & 17.48 & 28.18 & 26.78 & 25.62 & 15.91 & 26.68 & 17.19 & 19.33 & 9.87 & 22.59 \\
    % \midrule
    \specialrule{0.05em}{0pt}{0pt}
    %%%%%%%%%%%%%%%%%%%%%%%%%%%%%%%%%%%%%%
  \end{tabular}
  \end{adjustbox}
  \caption{BLEU results on the WMT24++ benchmark (document-level). Only models without WMT24++ training are included. Evaluation is conducted only on the intersection of language pairs supported by each baseline and our models. The first column (\# Langs) denotes the number of overlapping languages, followed by their distribution across resource tier (high/medium/low).
The symbol '-' indicates directions not supported by the baseline model.}
  \label{tab:8}
\end{table*}
\renewcommand{\arraystretch}{1}

\begin{table*}[htbp]
  \centering
  \begin{adjustbox}{max width=\textwidth}
  \begin{tabular}{ll}
    \toprule
    \textbf{Languages} & \textbf{Data Source} \\
    \midrule
    en & SlimPajama-6B \citep{cerebras2023slimpajama} \\
    \midrule
    zh & Skywork \citep{wei2023skywork} \\
    \midrule
    ar, th, vi & WanJuanSiLu \cite{yu2025wanjuansiluhighqualityopensourcewebtext} \\
    \midrule
    cs, fa, hi, de, fr, it, pt, es, el, uk, fi & \multirow{2}{*}{CulturaX\citep{nguyen2023culturaxcleanedenormousmultilingual}} \\
    sv, nb, da, ro, hu, sk, bg, nl, pl, tr  \\
    \midrule
    hr, sw & MADLAD-400 \citep{kudugunta2023madlad400} \\
    \midrule
    ne & CulturaX, MADLAD-400, MrBinit/Nepali \footnote{https://huggingface.co/datasets/MrBinit/nepali\_dataset\_text\_cleaned} \\
    \midrule
    bn, he, id, ms, az, kk, ps, ta, ur, uz, & CulturaX, MADLAD-400, Opus-Corpora\citep{zhang-etal-2020-improving}, \\
    am, jv & fineweb-2\citep{huggingfacefw_2024} \\
    \midrule
    \multirow{2}{*}{km, lo, my, tl} & CulturaX, MADLAD-400, fineweb-2, \\
    & Opus-Corpora, C4 \citep{2020t5}, OSCAR \footnote{https://oscar-project.github.io/documentation/versions/oscar-2301/} \\
    \midrule
    ja, ko, ru & WanJuanSiLu \citep{yu2025wanjuansiluhighqualityopensourcewebtext}\\
    \midrule
    ug, bo, mvf & In-house dataset \\
    \midrule
    yue & AlienKevin/yue\_and\_zh\_sentences \footnote{https://huggingface.co/datasets/AlienKevin/yue\_and\_zh\_sentences}\\
    \midrule
    si, te, mr, is, tg, ky, ka, hy & HPLT \citep{burchell2025expanded}, OSCAR, Wikipedia \\
    \bottomrule
  \end{tabular}
  \end{adjustbox}
  \caption{Monolingual data sources information for 60 languages.}
  \label{tab:lang_source}
\end{table*}

\begin{table*}
\label{tab:lang_info}
    %%%%%%%%%%%%%%%%%%%%%%%%%%%%%%%%%%%%%%%%%%%%%%%%%%%%%%%
    \begin{adjustbox}{max width=\textwidth, center}
    % \centering
    \begin{tabular}{p{2cm}p{3cm}p{3cm}p{3cm}p{2cm}}
    \toprule
    \textbf{ISO Code} & \textbf{Language} & \textbf{Script} & \textbf{Family} & \textbf{Resource} \\
    \specialrule{0.05em}{2pt}{0pt}
    \rowcolor{gray!20} \multicolumn{5}{l}{\textbf{13 High-resource Languages}} \\
    en & English & Latin & Indo-European & High\\
    ar & Arabic & Arabic & Afro-Asiatic & High\\
    es & Spanish & Latin & Indo-European & High\\
    de & German & Latin & Indo-European & High\\
    fr & French & Latin & Indo-European & High\\
    it & Italian & Latin & Indo-European & High\\
    ja & Japanese & Japanese & Japonic & High\\
    nl & Dutch & Latin & Indo-European & High\\
    pl & Polish & Latin & Indo-European & High\\
    pt & Portuguese & Latin & Indo-European & High\\
    ru & Russian & Cyrillic & Indo-European & High\\
    tr & Turkish & Latin & Turkic & High\\
    zh & Chinese & Han & Sino-Tibetan & High\\
    
    \specialrule{0.05em}{2pt}{0pt}
    \rowcolor{gray!20} \multicolumn{5}{l}{\textbf{18 Medium-resource Languages}} \\
    bg & Bulgarian & Cyrillic & Indo-European & Mid \\
    bn & Bengali & Bengali & Indo-European & Mid\\
    cs & Czech & Latin & Indo-European & Mid \\
    da & Danish & Latin & Indo-European & Mid \\
    el & Modern Greek & Greek & Indo-European & Mid\\
    fa & Persian & Arabic & Indo-European & Mid \\
    fi & Finnish & Latin & Uralic & Mid\\
    hi & Hindi & Devanagari & Indo-European & Mid \\
    hu & Hungarian & Latin & Uralic & Mid \\
    id & Indonesian & Latin & Austronesian & Mid \\
    ko & Korean & Hangul & Koreanic & Mid \\
    nb & Norwegian & Latin & Indo-European & Mid \\
    ro & Romanian & Latin & Indo-European & Mid \\
    sk & Slovak & Latin & Indo-European & Mid \\
    sv & Swedish & Latin & Indo-European & Mid \\
    th & Thai & Thai & Tai-Kadai & Mid \\
    uk & Ukrainian & Cyrillic & Indo-European & Mid \\
    vi & Vietnamese & Latin & Austroasiatic & Mid\\
    
    \specialrule{0.05em}{2pt}{0pt}
    \rowcolor{gray!20} \multicolumn{5}{l}{\textbf{29 Low-resource Languages}} \\
    am & Amharic & Ge’ez & Afro-Asiatic & Low\\
    az & Azerbaijani & Latin & Turkic & Low\\
    bo & Tibetan & Tibetan & Sino-Tibetan & Low \\
    he & Modern Hebrew & Hebrew & Afro-Asiatic & Low \\
    hr & Croatian & Latin & Indo-European & Low \\
    hy & Armenian & Armenian & Indo-European & Low \\
    is & Icelandic & Latin & Indo-European & Low \\
    jv & Javanese & Latin & Austronesian & Low \\
    ka & Georgian & Georgian & Kartvelian & Low \\
    kk & Kazakh & Cyrillic & Turkic & Low \\
    km & Central Khmer & Khmer & Austroasiatic & Low \\
    ky & Kirghiz & Cyrillic & Turkic & Low \\
    lo & Lao & Lao & Tai-Kadai & Low \\
    \bottomrule
    \end{tabular}
    \end{adjustbox}
\end{table*}

\begin{table*}[!ht]
\caption{Detailed information of 60 languages. The languages are grouped into categories based on their data ratios in the CulturaX \cite{nguyen-etal-2024-culturax}: High Resource (>1\%), Medium Resource (0.1\%–1\%], and Low Resource (≤0.1\%).}
% \label{tab:lang_info}
    \begin{adjustbox}{max width=\textwidth, center}
    % \centering
    \begin{tabular}{p{2cm}p{3cm}p{3cm}p{3cm}p{2cm}}
    \toprule
    \textbf{ISO Code} & \textbf{Language} & \textbf{Script} & \textbf{Family} & \textbf{Resource} \\
    \specialrule{0.05em}{2pt}{0pt}
    mvf & Inner Mongolian & Mongolian  & Mongolic & Low \\
    mr & Marathi & Devanagari & Indo-European & Low \\
    ms & Malay & Latin & Austronesian & Low \\
    my & Burmese & Myanmar & Sino-Tibetan & Low \\
    ne & Nepali & Devanagari & Indo-European & Low \\
    ps & Pashto & Arabic & Indo-European & Low \\
    si & Sinhala & Sinhala & Indo-European & Low \\
    sw & Swahili & Latin & Atlantic-Congo & Low \\
    ta & Tamil & Tamil & Dravidian & Low \\
    te & Telugu & Telugu & Dravidian & Low \\
    tg & Tajik & Cyrillic & Indo-European & Low \\
    tl & Tagalog & Latin & Austronesian & Low \\
    ug & Uighur & Arabic & Turkic & Low  \\
    ur & Urdu & Arabic & Indo-European & Low \\
    uz & Uzbek & Latin & Turkic & Low \\
    yue & Yue Chinese & Han & Sino-Tibetan & Low \\
    \bottomrule
    \end{tabular}
    \end{adjustbox}
\end{table*}
% \end{center}
% }

\begin{table*}[h]
\centering
% \begin{adjustbox}{max width=\textwidth}
% \setlength{\tabcolsep}{6pt} %% 默认列间距就是6pt
\begin{tabular}{p{2cm}p{2cm}p{2cm}p{3cm}p{2cm}p{1.6cm}}
\toprule
\textbf{ISO Code} & \textbf{Language} & \textbf{Script} & \textbf{Family} & \textbf{\begin{tabular}[c]{@{}c@{}}Auxiliary \\ Language\end{tabular}} & \textbf{ISO Code} \\
\midrule
es & Spanish & Latin & Indo-European & Portuguese & pt\\
de & German & Latin & Indo-European & Dutch & nl\\
fr & French & Latin & Indo-European & Italian & it\\
it & Italian & Latin & Indo-European & French & fr\\
nl & Dutch & Latin & Indo-European & German & de\\
pl & Polish & Latin & Indo-European & Czech & cs\\

bg & Bulgarian & Cyrillic & Indo-European & Russian & ru \\
cs & Czech & Latin & Indo-European & Polish & pl \\
da & Danish & Latin & Indo-European & Norwegian & nb \\
fa & Persian & Arabic & Indo-European & Arabic & ar \\
fi & Finnish & Latin & Uralic & Hungarian & hu\\
hi & Hindi & Devanagari & Indo-European & Bengali & bn\\
hu & Hungarian & Latin & Uralic & Finnish & fi\\
id & Indonesian & Latin & Austronesian & Dutch & nl\\
nb & Norwegian & Latin & Indo-European & Danish & da \\
ro & Romanian & Latin & Indo-European & Italian & it \\
sk & Slovak & Latin & Indo-European & Czech & cs \\
sv & Swedish & Latin & Indo-European & Norwegian & nb \\
uk & Ukrainian & Cyrillic & Indo-European & Russian & ru \\
vi & Vietnamese & Latin & Austroasiatic & French & fr \\

az & Azerbaijani & Latin & Turkic & Turkish & tr \\
hr & Croatian & Latin & Indo-European & Czech & cs \\
is & Icelandic & Latin & Indo-European & Danish & da \\
jv & Javanese & Latin & Austronesian & Indonesian & id \\
kk & Kazakh & Cyrillic & Turkic & Russian & ru \\
ky & Kirghiz & Cyrillic & Turkic & Russian & ru \\
lo & Lao & Lao & Tai-Kadai & Thai & th \\
mr & Marathi & Devanagari & Indo-European & Hindi & hi \\
ms & Malay & Latin & Austronesian & Indonesian & id \\
ne & Nepali & Devanagari & Indo-European & Hindi & hi \\
ps & Pashto & Arabic & Indo-European & Persian & fa \\
tg & Tajik & Cyrillic & Indo-European & Russian & ru \\
tl & Tagalog & Latin & Austronesian & Indonesian & id \\
ug & Uighur & Arabic & Turkic & Persian & fa \\
ur & Urdu & Arabic & Indo-European & Persian & fa \\
uz & Uzbek & Latin & Turkic & French & fr \\
\bottomrule
\end{tabular}
% \end{adjustbox}
\caption{The set of languages that utilize the Parallel Multilingual Prompting (PMP) method and their corresponding auxiliary languages.}
\label{tab:auxiliary}
\end{table*}

\input{appendices/new_en2x}
\input{appendices/new_x2en}
\input{appendices/new_zh2x}

\input{appendices/new_x2zh}

%% file: appendices/Translation_Error.tex
\definecolor{errorcolor}{RGB}{191,  1, 53}
\newcommand{\mybold}[1]{{\bfseries\color{errorcolor}#1}}

\begin{table*}[htbp]
    \centering
    \caption{Translation error cases: hallucinations and fabricated outputs}
    \label{tab:error_case}
    \renewcommand{\arraystretch}{1} 
    \begin{tabularx}{\textwidth}{lX}
        \toprule
        \textbf{Category} & \textbf{Content} \\
        \midrule
        
        \textbf{Source} & 他补充道：“我们现在有 4 个月大没有糖尿病的老鼠，但它们 \mybold{曾经得过该病}。” \\
        \addlinespace
        
        \textbf{Reference} & "We now have 4-month-old mice that are non-diabetic that \mybold{used to be diabetic}," he added.  \\
        \addlinespace
        
        \textbf{Prediction} & He added: "We now have four-month-old mice that \mybold{have never had diabetes but were given the disease}." \\
        
        \bottomrule
    \end{tabularx}
\end{table*}

%% file: appendices/new_en2x.tex
% Table 1: en→x
\renewcommand\arraystretch{0.9}
\begin{table*}[htbp]
    \begin{adjustbox}{max width=1.0\textwidth, center=\textwidth}
    \setlength{\tabcolsep}{15pt}
    \begin{tabular}{lcccccccc}
        % \hline
        \toprule
        \multirow{2.5}{*}{\textbf{Direction}}&\multicolumn{4}{c}{\textbf{COMET}}&\multicolumn{4}{c}{\textbf{BLEU}}\\
        \cmidrule(lr){2-5} \cmidrule(lr){6-9}
        ~ &\textbf{0.6B}&\textbf{1.7B}&\textbf{4B}&\textbf{8B} &\textbf{0.6B}&\textbf{1.7B}&\textbf{4B}&\textbf{8B}\\
        \hline
        \rowcolor{gray!20} \multicolumn{9}{l}{\textbf{13 High-resource Languages}} \\
        en→ar  &82.97  &85.51  &87.01  &87.14  &17.64  &19.45  &22.54  &25.09  \\
        en→es  &84.77  &86.67  &87.32  &87.40  &24.82  &27.14  &29.04  &29.39  \\
        en→de  &83.58  &87.66  &88.54  &88.87  &28.92  &35.39  &37.74  &38.79  \\
        en→fr  &85.70  &88.10  &88.70  &89.01  &40.79  &45.41  &47.42  &49.50  \\
        en→it  &86.17  &88.58  &89.40  &89.49  &25.70  &27.99  &30.22  &31.21  \\
        en→ja  &90.09  &91.61  &92.26  &92.38  &30.69  &35.35  &37.24  &38.35  \\
        en→nl  &84.75  &87.72  &88.65  &88.78  &22.32  &24.82  &26.11  &27.42  \\
        en→pl  &84.48  &88.86  &90.04  &90.54  &15.58  &19.31  &21.37  &22.33  \\
        en→pt  &88.03  &89.78  &90.32  &90.41  &42.78  &46.27  &48.32  &49.37  \\
        en→ru  &86.49  &89.22  &90.37  &90.47  &24.92  &28.20  &30.73  &31.66  \\
        en→tr  &86.72  &89.74  &90.78  &90.90  &20.36  &23.86  &25.59  &27.55  \\
        en→zh  &87.92  &89.49  &89.82  &89.84  &44.12  &48.03  &49.21  &49.34  \\
        \hline
        \rowcolor{gray!20} \multicolumn{9}{l}{\textbf{18 Medium-resource Languages}} \\
        en→bg  &87.90  &90.74  &91.39  &91.58  &31.84  &35.90  &37.52  &38.85  \\
        en→bn  &83.20  &86.81  &87.85  &87.60  &13.47  &15.01  &15.91  &16.99  \\
        en→cs  &87.19  &91.04  &92.05  &92.18  &25.48  &28.95  &31.15  &32.30  \\
        en→da  &88.27  &90.60  &91.65  &91.76  &38.27  &41.34  &44.11  &45.04  \\
        en→el  &84.59  &88.68  &89.71  &90.08  &19.39  &22.54  &24.99  &26.10  \\
        en→fa  &82.58  &86.70  &88.31  &88.51  &19.56  &22.53  &24.64  &25.85  \\
        en→fi  &85.85  &90.98  &92.67  &92.83  &15.25  &19.18  &22.85  &23.77  \\
        en→hi  &76.81  &80.45  &81.85  &82.28  &24.11  &27.00  &28.85  &30.45  \\
        en→hu  &83.81  &88.80  &90.26  &90.28  &17.04  &20.48  &23.59  &24.14  \\
        en→id  &90.62  &91.93  &92.38  &92.47  &43.23  &45.00  &45.71  &47.10  \\
        en→ko  &87.55  &89.38  &90.47  &90.47  &26.61  &28.17  &29.38  &31.36  \\
        en→nb  &87.89  &89.91  &90.64  &90.77  &28.57  &30.39  &31.46  &31.87  \\
        en→ro  &86.68  &90.20  &91.36  &91.55  &32.33  &35.74  &38.28  &40.00  \\
        en→sk  &85.90  &90.22  &91.44  &91.86  &25.60  &28.56  &31.41  &32.84  \\
        en→sv  &87.40  &90.52  &91.51  &91.63  &36.37  &39.99  &42.74  &43.44  \\
        en→th  &85.83  &88.63  &89.79  &89.86  &11.56  &11.82  &14.14  &14.94  \\
        en→uk  &85.08  &89.50  &90.64  &91.00  &20.86  &25.40  &28.64  &30.21  \\
        en→vi  &88.25  &89.65  &90.22  &90.31  &39.10  &40.82  &41.88  &42.74  \\
        \hline
        \rowcolor{gray!20} \multicolumn{9}{l}{\textbf{29 Low-resource Languages}} \\
        en→am  &79.76  &86.81  &88.40  &88.67  &5.38  &8.57  &10.77  &11.57  \\
        en→az  &84.31  &88.17  &89.25  &89.15  &11.80  &13.04  &14.23  &14.38  \\
        en→bo  &87.37  &91.51  &92.82  &93.09  &1.71  &3.63  &3.26  &2.84  \\
        en→he  &81.79  &83.31  &87.55  &87.80  &17.16  &20.37  &24.68  &27.15  \\
        en→hr  &86.02  &90.06  &91.56  &91.58  &22.48  &25.60  &28.98  &29.62  \\
        en→hy  &86.42  &89.90  &91.10  &91.11  &17.23  &20.13  &22.33  &23.10  \\
        en→is  &81.43  &85.65  &86.91  &87.38  &18.77  &21.95  &23.75  &24.53  \\
        en→jv  &85.89  &87.89  &88.37  &88.41  &23.92  &27.22  &28.74  &29.58  \\
        en→ka  &81.89  &86.89  &89.02  &89.17  &10.89  &12.22  &14.84  &14.90  \\
        en→kk  &88.46  &91.05  &91.53  &91.69  &18.53  &21.52  &23.06  &23.76  \\
        en→km  &79.32  &83.38  &85.04  &85.11  &6.52  &7.72  &7.50  &7.49  \\
        en→ky  &85.84  &89.11  &89.98  &89.84  &11.35  &13.58  &14.45  &14.73  \\
        en→lo  &80.52  &85.09  &86.81  &86.78  &11.57  &13.64  &14.31  &14.42  \\
        en→mvf  &95.25  &95.22  &95.48  &95.55  &15.12  &11.74  &15.17  &16.67  \\
        en→mr  &70.80  &75.00  &76.45  &76.67  &11.20  &14.13  &15.54  &16.72  \\
        en→ms  &88.98  &90.23  &90.62  &90.70  &39.25  &40.67  &41.05  &41.95  \\
        en→my  &77.47  &86.78  &88.79  &88.97  &2.72  &3.36  &4.28  &4.51  \\
        en→ne  &81.45  &84.14  &85.23  &85.22  &15.79  &18.21  &19.45  &20.49  \\
        en→ps  &75.67  &79.49  &80.25  &80.15  &9.36  &11.06  &12.04  &12.03  \\
        en→si  &82.90  &88.20  &90.29  &90.64  &10.33  &12.89  &15.50  &17.16  \\
        en→sw  &82.88  &85.47  &86.08  &86.35  &30.52  &33.45  &35.56  &36.72  \\
        en→ta  &85.42  &89.05  &90.27  &90.35  &10.79  &13.07  &14.29  &15.33  \\
        en→te  &83.10  &86.59  &87.49  &87.59  &15.33  &17.57  &19.29  &20.54  \\
        en→tg  &75.64  &77.25  &77.68  &77.81  &15.19  &18.64  &20.73  &21.87  \\
        en→tl  &83.16  &85.42  &86.37  &86.18  &32.19  &35.54  &37.34  &38.30  \\
        en→ug  &84.05  &87.82  &88.93  &88.68  &12.13  &14.52  &16.70  &18.01  \\
        en→ur  &78.67  &82.87  &84.23  &84.08  &17.40  &19.49  &21.38  &22.16  \\
        en→uz  &88.52  &90.74  &91.45  &91.32  &15.18  &18.21  &20.46  &20.81  \\
        en→yue  &85.88  &88.85  &89.63  &89.85  &5.02  &9.32  &5.58  &8.32  \\
        % \hline
        \bottomrule
    \end{tabular}
    \end{adjustbox}
    \caption{COMET-22 and SacreBLEU scores of LMT on the FLORES-200 devtest set (En $\to$ X).}
\end{table*}
\renewcommand\arraystretch{1.0}

%% file: appendices/new_x2en.tex
% Table 2: x→en
\renewcommand\arraystretch{0.9}
\begin{table*}[htbp]
    \begin{adjustbox}{max width=1.0\textwidth, center=\textwidth}
    \setlength{\tabcolsep}{15pt}
    \begin{tabular}{lcccccccc}
        % \hline
        \toprule
        \multirow{2.5}{*}{\textbf{Direction}}&\multicolumn{4}{c}{\textbf{COMET}}&\multicolumn{4}{c}{\textbf{BLEU}}\\
        \cmidrule(lr){2-5} \cmidrule(lr){6-9}
        ~ &\textbf{0.6B}&\textbf{1.7B}&\textbf{4B}&\textbf{8B} &\textbf{0.6B}&\textbf{1.7B}&\textbf{4B}&\textbf{8B}\\
        \hline
        \rowcolor{gray!20} \multicolumn{9}{l}{\textbf{13 High-resource Languages}} \\
        ar→en  &84.53  &87.15  &87.78  &87.97  &33.56  &38.83  &40.93  &41.40  \\
        es→en  &86.38  &87.41  &87.85  &87.94  &29.17  &31.27  &33.64  &33.57  \\
        de→en  &88.14  &89.38  &89.59  &89.52  &39.99  &43.33  &44.45  &43.99  \\
        fr→en  &88.43  &89.34  &89.48  &89.49  &41.93  &44.98  &45.34  &45.71  \\
        it→en  &87.09  &88.14  &88.60  &88.59  &31.12  &33.97  &35.88  &35.95  \\
        ja→en  &86.85  &88.30  &88.71  &88.82  &25.51  &28.35  &30.40  &31.09  \\
        nl→en  &86.27  &87.70  &87.84  &87.94  &29.92  &32.36  &33.05  &33.56  \\
        pl→en  &84.47  &86.36  &86.74  &86.72  &27.50  &30.83  &31.93  &31.94  \\
        pt→en  &88.54  &89.53  &89.98  &89.96  &45.49  &48.63  &50.56  &50.51  \\
        ru→en  &85.50  &86.81  &87.21  &87.36  &31.91  &35.56  &37.70  &38.03  \\
        tr→en  &87.12  &89.28  &89.76  &89.83  &32.26  &36.79  &39.18  &39.31  \\
        zh→en  &86.73  &87.79  &87.96  &87.87  &28.62  &30.95  &32.59  &32.88  \\
        \hline
        \rowcolor{gray!20} \multicolumn{9}{l}{\textbf{18 Medium-resource Languages}} \\
        bg→en  &86.72  &88.05  &88.47  &88.40  &36.89  &40.82  &41.85  &41.68  \\
        bn→en  &85.21  &88.04  &88.98  &89.25  &25.12  &32.25  &34.61  &35.73  \\
        cs→en  &86.78  &88.48  &88.98  &88.88  &35.38  &39.52  &41.44  &41.28  \\
        da→en  &88.86  &90.15  &90.36  &90.27  &44.61  &48.59  &48.66  &47.73  \\
        el→en  &85.09  &87.49  &88.12  &88.14  &30.94  &35.74  &37.86  &38.17  \\
        fa→en  &85.32  &87.87  &88.48  &88.65  &30.48  &35.40  &37.34  &38.10  \\
        fi→en  &86.46  &89.36  &90.09  &90.23  &27.84  &33.16  &34.76  &35.45  \\
        hi→en  &86.97  &89.16  &89.94  &89.99  &31.72  &38.53  &40.53  &40.44  \\
        hu→en  &86.30  &88.36  &88.91  &88.93  &30.87  &35.52  &37.16  &37.17  \\
        id→en  &88.61  &89.72  &89.93  &89.95  &41.19  &44.60  &45.69  &45.56  \\
        ko→en  &86.62  &88.31  &88.68  &88.61  &25.94  &30.12  &30.95  &31.66  \\
        nb→en  &87.57  &88.96  &89.24  &89.29  &41.13  &43.75  &44.30  &44.40  \\
        ro→en  &88.00  &89.42  &89.59  &89.66  &39.91  &43.51  &44.88  &44.81  \\
        sk→en  &86.32  &88.24  &88.64  &88.51  &34.59  &39.32  &40.64  &40.35  \\
        sv→en  &88.52  &90.04  &90.27  &90.26  &44.68  &48.14  &48.82  &48.00  \\
        th→en  &86.78  &88.62  &89.13  &89.01  &28.22  &32.63  &34.02  &34.46  \\
        uk→en  &85.55  &87.35  &87.71  &87.64  &35.01  &39.42  &40.15  &40.47  \\
        vi→en  &86.70  &87.83  &88.21  &88.15  &35.14  &37.47  &38.37  &38.51  \\
        \hline
        \rowcolor{gray!20} \multicolumn{9}{l}{\textbf{29 Low-resource Languages}} \\
        am→en  &77.78  &83.41  &86.36  &87.20  &15.76  &25.52  &31.42  &33.47  \\
        az→en  &84.26  &86.46  &87.42  &87.45  &19.75  &23.76  &26.12  &26.73  \\
        bo→en  &64.15  &70.09  &72.90  &74.04  &6.20  &11.59  &15.67  &17.72  \\
        he→en  &84.98  &87.72  &88.58  &88.65  &35.83  &41.79  &44.13  &44.58  \\
        hr→en  &85.83  &88.01  &88.41  &88.37  &33.44  &38.05  &39.41  &39.15  \\
        hy→en  &85.31  &87.82  &88.75  &88.98  &30.29  &36.15  &38.69  &40.41  \\
        is→en  &82.28  &85.90  &87.04  &87.41  &28.44  &33.98  &36.39  &37.38  \\
        jv→en  &81.57  &84.96  &86.30  &86.53  &32.90  &38.37  &41.37  &41.94  \\
        ka→en  &83.54  &86.35  &87.29  &87.62  &22.45  &26.80  &29.33  &30.21  \\
        kk→en  &84.96  &87.67  &88.29  &88.57  &27.53  &32.83  &34.90  &35.83  \\
        km→en  &83.37  &86.26  &87.56  &88.02  &23.47  &30.17  &34.01  &35.15  \\
        ky→en  &83.17  &85.84  &86.66  &86.65  &20.17  &23.50  &25.54  &26.25  \\
        lo→en  &83.32  &86.02  &87.82  &88.31  &25.63  &31.88  &36.87  &38.06  \\
        mvf→en  &69.36  &74.64  &76.58  &77.86  &10.00  &14.22  &18.04  &20.45  \\
        mr→en  &84.96  &87.71  &88.73  &88.73  &27.81  &33.60  &36.46  &37.52  \\
        ms→en  &88.02  &89.28  &89.59  &89.65  &41.94  &45.06  &45.99  &46.28  \\
        my→en  &80.74  &84.80  &86.62  &86.88  &17.59  &23.72  &26.20  &27.78  \\
        ne→en  &87.81  &89.95  &90.66  &90.90  &30.23  &37.25  &39.12  &40.79  \\
        ps→en  &80.52  &83.94  &85.41  &85.76  &24.03  &29.30  &32.95  &33.63  \\
        si→en  &81.72  &85.36  &88.44  &89.00  &19.28  &26.12  &32.61  &34.32  \\
        sw→en  &81.64  &84.74  &86.26  &86.37  &34.29  &39.21  &42.08  &43.14  \\
        ta→en  &82.22  &86.08  &87.38  &87.70  &22.40  &28.89  &31.98  &33.11  \\
        te→en  &84.09  &87.62  &88.78  &89.10  &26.63  &35.44  &37.64  &39.50  \\
        tg→en  &73.45  &77.79  &79.12  &79.49  &26.06  &32.20  &34.85  &36.21  \\
        tl→en  &84.73  &87.21  &88.15  &88.39  &38.80  &44.41  &47.59  &48.05  \\
        ug→en  &83.68  &86.76  &87.74  &87.84  &21.31  &25.96  &28.41  &28.83  \\
        ur→en  &83.64  &87.08  &87.99  &88.23  &26.55  &33.38  &35.50  &36.67  \\
        uz→en  &85.05  &87.69  &88.42  &88.57  &28.58  &34.00  &35.75  &36.32  \\
        yue→en  &86.53  &87.81  &88.13  &88.06  &28.84  &32.09  &32.60  &33.58  \\
        % \hline
        \bottomrule
    \end{tabular}
    \end{adjustbox}
    \caption{COMET-22 and SacreBLEU scores of LMT on the FLORES-200 devtest set (X $\to$ En).}
\end{table*}
\renewcommand\arraystretch{1.0}

%% file: appendices/new_zh2x.tex
% Table 4: zh→x
\renewcommand\arraystretch{0.9}
\begin{table*}[htbp]
    \begin{adjustbox}{max width=1.0\textwidth, center=\textwidth}
    \setlength{\tabcolsep}{15pt}
    \begin{tabular}{lcccccccc}
        % \hline
        \toprule
        \multirow{2.5}{*}{\textbf{Direction}}&\multicolumn{4}{c}{\textbf{COMET}}&\multicolumn{4}{c}{\textbf{BLEU}}\\
        \cmidrule(lr){2-5} \cmidrule(lr){6-9}
        ~ &\textbf{0.6B}&\textbf{1.7B}&\textbf{4B}&\textbf{8B} &\textbf{0.6B}&\textbf{1.7B}&\textbf{4B}&\textbf{8B}\\
        \hline
        \rowcolor{gray!20} \multicolumn{9}{l}{\textbf{13 High-resource Languages}} \\
        zh→ar  &79.71  &83.02  &84.27  &84.31  &8.71  &10.67  &13.14  &13.64  \\
        zh→en  &86.73  &87.79  &87.96  &87.87  &28.62  &30.95  &32.59  &32.88  \\
        zh→es  &82.67  &85.05  &85.67  &86.03  &16.32  &19.01  &20.04  &21.03  \\
        zh→de  &80.41  &84.75  &85.58  &86.10  &14.97  &18.62  &21.15  &22.57  \\
        zh→fr  &82.06  &84.86  &85.64  &85.80  &21.16  &25.07  &27.75  &29.36  \\
        zh→it  &83.75  &86.70  &87.61  &87.71  &15.70  &18.60  &20.35  &21.48  \\
        zh→ja  &88.87  &90.71  &91.20  &91.21  &24.08  &27.48  &29.90  &30.81  \\
        zh→nl  &81.67  &85.47  &86.44  &86.66  &13.11  &15.77  &17.65  &18.91  \\
        zh→pl  &83.08  &87.95  &89.16  &89.41  &9.86  &12.36  &14.86  &15.16  \\
        zh→pt  &84.89  &86.97  &87.46  &87.75  &21.17  &24.49  &26.04  &27.65  \\
        zh→ru  &84.76  &88.19  &88.94  &89.24  &14.21  &17.14  &18.67  &20.11  \\
        zh→tr  &81.09  &85.33  &86.51  &86.80  &10.47  &12.93  &14.87  &16.15  \\
        \hline
        \rowcolor{gray!20} \multicolumn{9}{l}{\textbf{18 Medium-resource Languages}} \\
        zh→bg  &84.63  &88.06  &89.24  &89.39  &17.14  &19.66  &21.96  &23.17  \\
        zh→bn  &78.06  &82.14  &83.52  &83.57  &6.72  &7.98  &9.16  &10.22  \\
        zh→cs  &84.72  &89.18  &90.32  &90.65  &13.30  &17.02  &18.97  &19.98  \\
        zh→da  &84.95  &87.85  &88.70  &88.97  &18.89  &21.87  &24.13  &25.02  \\
        zh→el  &81.00  &85.66  &87.02  &87.53  &10.69  &12.65  &15.54  &16.67  \\
        zh→fa  &79.75  &84.04  &85.86  &85.99  &11.50  &13.87  &16.00  &16.39  \\
        zh→fi  &81.87  &87.79  &89.79  &90.17  &8.83  &10.96  &14.32  &15.12  \\
        zh→hi  &69.07  &73.82  &75.37  &75.50  &12.42  &15.62  &17.77  &18.27  \\
        zh→hu  &79.62  &85.75  &87.26  &87.52  &9.78  &12.58  &15.56  &15.94  \\
        zh→id  &87.07  &88.80  &89.27  &89.35  &23.24  &25.21  &27.07  &27.71  \\
        zh→ko  &85.14  &87.68  &88.49  &88.56  &18.88  &21.48  &23.20  &24.20  \\
        zh→nb  &84.44  &87.44  &88.22  &88.26  &13.90  &16.60  &17.92  &18.74  \\
        zh→ro  &83.24  &87.03  &88.26  &88.51  &17.83  &21.19  &23.41  &24.54  \\
        zh→sk  &83.23  &88.15  &89.65  &89.65  &13.10  &15.49  &18.31  &19.44  \\
        zh→sv  &84.32  &87.64  &88.70  &88.89  &17.00  &20.12  &22.68  &23.97  \\
        zh→th  &83.82  &86.58  &87.42  &87.56  &8.20  &9.17  &10.46  &11.87  \\
        zh→uk  &82.86  &87.74  &89.13  &89.42  &10.96  &14.72  &16.84  &17.96  \\
        zh→vi  &86.97  &88.57  &89.06  &89.12  &27.63  &29.38  &31.07  &32.32  \\
        \hline
        \rowcolor{gray!20} \multicolumn{9}{l}{\textbf{29 Low-resource Languages}} \\
        zh→am  &74.21  &81.93  &84.53  &84.54  &2.61  &4.24  &5.60  &6.33  \\
        zh→az  &80.11  &85.27  &86.28  &86.43  &7.95  &9.61  &10.57  &10.75  \\
        zh→bo  &88.72  &92.10  &93.26  &93.03  &1.61  &2.39  &2.12  &2.17  \\
        zh→he  &78.05  &81.34  &84.69  &84.41  &7.97  &10.43  &14.13  &14.53  \\
        zh→hr  &83.28  &88.23  &89.83  &90.14  &12.13  &14.94  &18.07  &18.99  \\
        zh→hy  &82.24  &86.51  &88.01  &88.19  &9.16  &11.54  &13.25  &13.18  \\
        zh→is  &78.16  &82.82  &84.73  &84.71  &10.50  &12.47  &15.06  &15.34  \\
        zh→jv  &81.64  &84.92  &85.38  &85.35  &11.06  &14.80  &16.08  &16.47  \\
        zh→ka  &77.88  &84.30  &86.20  &86.62  &6.61  &8.30  &10.02  &11.27  \\
        zh→kk  &84.72  &87.86  &88.69  &88.82  &10.15  &12.27  &13.89  &14.60  \\
        zh→km  &75.69  &80.56  &82.11  &82.22  &5.01  &5.41  &6.41  &6.52  \\
        zh→ky  &82.14  &86.74  &87.36  &87.67  &7.62  &9.53  &10.47  &11.11  \\
        zh→lo  &76.58  &81.78  &83.74  &83.86  &7.26  &9.42  &10.01  &10.06  \\
        zh→mvf  &96.15  &95.96  &96.20  &96.25  &25.34  &25.23  &31.97  &35.50  \\
        zh→mr  &63.03  &68.39  &70.03  &70.45  &6.15  &7.81  &9.02  &9.65  \\
        zh→ms  &85.16  &86.77  &87.09  &87.21  &20.51  &22.09  &23.22  &24.06  \\
        zh→my  &71.97  &83.07  &85.30  &85.61  &1.60  &2.40  &2.87  &2.88  \\
        zh→ne  &73.18  &77.19  &78.04  &78.19  &7.17  &8.69  &9.88  &10.19  \\
        zh→ps  &70.92  &76.46  &77.23  &77.25  &5.10  &6.65  &8.09  &7.94  \\
        zh→si  &78.49  &85.07  &87.50  &87.84  &5.71  &7.68  &9.68  &10.54  \\
        zh→sw  &78.23  &81.45  &82.27  &82.63  &13.87  &16.80  &18.80  &19.85  \\
        zh→ta  &80.41  &85.38  &86.49  &86.78  &5.95  &7.35  &8.67  &8.84  \\
        zh→te  &75.90  &81.20  &82.67  &82.92  &7.20  &9.11  &10.35  &11.45  \\
        zh→tg  &73.88  &75.87  &76.64  &76.33  &8.47  &11.36  &13.42  &14.02  \\
        zh→tl  &78.69  &81.84  &82.61  &82.57  &16.05  &19.96  &21.22  &22.65  \\
        zh→ug  &80.12  &84.64  &85.59  &85.49  &9.20  &11.29  &12.73  &13.86  \\
        zh→ur  &73.01  &78.45  &79.89  &80.26  &9.16  &11.82  &13.22  &14.51  \\
        zh→uz  &84.90  &87.85  &88.65  &88.62  &8.46  &10.58  &12.11  &12.59  \\
        zh→yue  &90.19  &91.55  &91.74  &91.84  &5.52  &8.73  &4.76  &7.64  \\
        % \hline
        \bottomrule
    \end{tabular}
    \end{adjustbox}
    \caption{COMET-22 and SacreBLEU scores of LMT on the FLORES-200 devtest set (Zh $\to$ X).}
\end{table*}
\renewcommand\arraystretch{1.0}

%% file: appendices/new_x2zh.tex
% Table 3: x→zh
\renewcommand\arraystretch{0.9}
\begin{table*}[htbp]
    \begin{adjustbox}{max width=1.0\textwidth, center=\textwidth}
    \setlength{\tabcolsep}{15pt}
    \begin{tabular}{lcccccccc}
        % \hline
        \toprule
        \multirow{2.5}{*}{\textbf{Direction}}&\multicolumn{4}{c}{\textbf{COMET}}&\multicolumn{4}{c}{\textbf{BLEU}}\\
        \cmidrule(lr){2-5} \cmidrule(lr){6-9}
        ~ &\textbf{0.6B}&\textbf{1.7B}&\textbf{4B}&\textbf{8B} &\textbf{0.6B}&\textbf{1.7B}&\textbf{4B}&\textbf{8B}\\
        \hline
        \rowcolor{gray!20} \multicolumn{9}{l}{\textbf{13 High-resource Languages}} \\
        ar→zh  &82.33  &85.82  &86.77  &86.89  &31.79  &38.23  &40.56  &41.53  \\
        en→zh  &87.92  &89.49  &89.82  &89.84  &44.12  &48.03  &49.21  &49.34  \\
        es→zh  &86.04  &87.73  &88.33  &88.32  &34.09  &37.72  &38.84  &39.49  \\
        de→zh  &85.80  &87.94  &88.43  &88.46  &36.81  &41.25  &42.37  &42.21  \\
        fr→zh  &86.55  &87.94  &88.49  &88.36  &37.54  &41.60  &42.80  &42.83  \\
        it→zh  &86.18  &87.93  &88.49  &88.41  &34.67  &38.93  &40.30  &40.79  \\
        ja→zh  &86.80  &88.73  &89.36  &89.16  &31.92  &36.96  &38.52  &38.68  \\
        nl→zh  &84.78  &86.89  &87.52  &87.53  &32.49  &36.40  &38.20  &38.10  \\
        pl→zh  &83.95  &86.61  &87.19  &87.28  &31.89  &36.34  &38.01  &38.38  \\
        pt→zh  &86.46  &88.35  &88.86  &88.83  &37.45  &41.65  &42.81  &43.32  \\
        ru→zh  &84.87  &86.83  &87.42  &87.40  &35.11  &39.06  &40.93  &40.64  \\
        tr→zh  &83.99  &86.68  &87.58  &87.56  &32.56  &38.48  &40.65  &40.64  \\
        \hline
        \rowcolor{gray!20} \multicolumn{9}{l}{\textbf{18 Medium-resource Languages}} \\
        bg→zh  &84.40  &86.94  &87.59  &87.57  &35.67  &40.08  &41.89  &41.68  \\
        bn→zh  &81.62  &85.77  &87.13  &87.45  &27.02  &34.16  &36.67  &37.84  \\
        cs→zh  &85.06  &87.33  &88.01  &87.82  &34.77  &39.89  &41.41  &41.42  \\
        da→zh  &85.99  &88.27  &88.98  &88.69  &37.27  &42.14  &43.40  &42.99  \\
        el→zh  &82.17  &85.81  &86.94  &86.91  &30.55  &36.86  &39.04  &39.39  \\
        fa→zh  &83.02  &86.58  &87.53  &87.47  &30.68  &37.07  &38.64  &39.38  \\
        fi→zh  &83.35  &87.26  &88.12  &88.25  &30.90  &37.21  &39.83  &39.49  \\
        hi→zh  &82.65  &86.27  &87.32  &87.67  &30.19  &36.55  &39.01  &39.84  \\
        hu→zh  &83.67  &86.82  &87.61  &87.48  &32.40  &38.49  &40.41  &39.98  \\
        id→zh  &85.98  &87.72  &88.25  &88.21  &36.96  &41.91  &42.57  &42.99  \\
        ko→zh  &85.44  &87.71  &88.28  &88.43  &32.16  &36.68  &38.59  &39.34  \\
        nb→zh  &85.26  &87.48  &88.20  &88.26  &34.91  &39.68  &41.22  &41.07  \\
        ro→zh  &84.94  &87.45  &88.06  &88.02  &36.41  &41.45  &42.90  &42.70  \\
        sk→zh  &84.61  &87.27  &87.77  &87.75  &34.06  &39.29  &40.85  &41.18  \\
        sv→zh  &85.93  &88.18  &88.90  &88.85  &36.53  &41.38  &42.79  &42.89  \\
        th→zh  &85.72  &88.20  &88.78  &88.70  &32.22  &37.83  &39.51  &39.52  \\
        uk→zh  &83.86  &86.61  &87.32  &87.19  &34.54  &39.76  &41.31  &41.11  \\
        vi→zh  &86.53  &88.24  &88.61  &88.46  &35.73  &40.02  &40.87  &41.02  \\
        \hline
        \rowcolor{gray!20} \multicolumn{9}{l}{\textbf{29 Low-resource Languages}} \\
        am→zh  &73.26  &80.52  &83.92  &84.74  &16.53  &26.33  &31.58  &33.71  \\
        az→zh  &82.36  &85.47  &86.37  &86.44  &27.51  &31.84  &34.02  &34.52  \\
        bo→zh  &64.72  &72.11  &74.55  &76.06  &7.84  &17.05  &21.15  &23.54  \\
        he→zh  &82.45  &86.12  &87.27  &87.39  &32.04  &38.68  &41.32  &41.76  \\
        hr→zh  &84.01  &87.16  &87.95  &87.91  &33.55  &38.71  &40.51  &40.49  \\
        hy→zh  &81.99  &85.80  &87.24  &87.32  &30.52  &36.90  &39.66  &40.35  \\
        is→zh  &80.51  &85.02  &86.20  &86.30  &28.44  &34.82  &37.69  &38.34  \\
        jv→zh  &79.12  &83.44  &85.00  &85.06  &29.27  &35.64  &37.69  &38.54  \\
        ka→zh  &80.88  &85.28  &86.86  &86.99  &25.37  &33.32  &36.54  &37.23  \\
        kk→zh  &82.87  &85.96  &87.03  &87.13  &30.63  &36.15  &39.02  &38.87  \\
        km→zh  &81.26  &84.94  &86.58  &86.73  &26.41  &32.17  &35.44  &36.42  \\
        ky→zh  &81.23  &85.01  &86.24  &86.14  &25.83  &31.01  &33.91  &34.09  \\
        lo→zh  &80.42  &84.79  &86.53  &87.11  &25.25  &32.50  &36.13  &38.08  \\
        mvf→zh  &77.98  &81.50  &83.95  &85.12  &24.01  &46.70  &54.35  &58.64  \\
        mr→zh  &80.23  &85.17  &86.60  &86.63  &27.00  &35.08  &37.42  &37.95  \\
        ms→zh  &84.87  &87.04  &87.73  &87.86  &36.14  &40.54  &41.99  &42.03  \\
        my→zh  &76.46  &83.03  &85.10  &85.40  &14.78  &26.54  &30.40  &31.71  \\
        ne→zh  &83.16  &86.41  &87.78  &88.13  &29.26  &35.60  &37.99  &39.22  \\
        ps→zh  &77.75  &82.73  &84.59  &84.85  &23.07  &30.92  &34.43  &35.06  \\
        si→zh  &78.07  &83.39  &86.90  &87.38  &19.46  &27.73  &35.46  &36.24  \\
        sw→zh  &78.09  &83.09  &84.73  &85.03  &27.18  &35.13  &37.31  &38.63  \\
        ta→zh  &78.56  &83.57  &85.34  &85.67  &22.72  &31.58  &34.49  &35.40  \\
        te→zh  &78.81  &84.35  &86.10  &86.57  &24.11  &34.04  &37.07  &37.82  \\
        tg→zh  &73.04  &77.70  &79.72  &79.93  &27.95  &34.58  &37.93  &38.51  \\
        tl→zh  &81.68  &85.14  &86.36  &86.39  &32.58  &38.84  &41.48  &41.47  \\
        ug→zh  &81.64  &85.73  &86.90  &87.10  &28.28  &34.81  &37.32  &37.18  \\
        ur→zh  &80.30  &84.70  &86.38  &86.49  &25.39  &33.71  &37.21  &37.50  \\
        uz→zh  &82.42  &85.95  &87.17  &87.16  &30.56  &37.25  &39.21  &39.38  \\
        yue→zh  &90.67  &91.42  &91.49  &91.39  &43.02  &46.05  &45.82  &45.25  \\
        % \hline
        \bottomrule
    \end{tabular}
    \end{adjustbox}
    \caption{COMET-22 and SacreBLEU scores of LMT on the FLORES-200 devtest set (X $\to$ Zh).}
\end{table*}
\renewcommand\arraystretch{1.0}